\documentclass[10pt,twocolumn,letterpaper]{article}

\usepackage[pagenumbers]{vjlm}

\usepackage{colortbl}
\usepackage{multirow}
\usepackage{amsthm}
\usepackage{tikz}
\usepackage{pgfplots}
\pgfplotsset{compat=1.18}
\usetikzlibrary{arrows.meta,positioning}
\definecolor{cvj}{RGB}{42,120,214}     
\definecolor{cvjl}{RGB}{27,175,122}    
\definecolor{csg}{RGB}{235,104,52}     
\definecolor{ours}{RGB}{232,240,250}   
\theoremstyle{definition}

\newtheorem{proposition}{Proposition}
\usepackage{xspace}
\newcommand{\ours}{V-JEPA-LM\xspace}
\usepackage[scaled=0.92]{helvet}   
\definecolor{cpA}{RGB}{88,98,118}     
\definecolor{cpB}{RGB}{108,78,198}    
\definecolor{cpC}{RGB}{52,132,86}     
\usepackage{array}

\usepackage{placeins}

\definecolor{vjlmblue}{rgb}{0.21,0.49,0.74}
\usepackage[pagebackref,breaklinks,colorlinks,allcolors=vjlmblue]{hyperref}

\def\confName{}
\def\confYear{2027}

\title{V-JEPA-LM: Video-Language Models That Read Dynamics\\Beyond Static, Text-Aligned Representations}

\author{Keuntae Kim\qquad Dongjun Kim\qquad Yong Suk Choi\thanks{Corresponding author: \texttt{cys@hanyang.ac.kr}}\\
Department of Computer Science, Hanyang University, Seoul, Republic of Korea\\
{\tt\small \{ktkpv94, cys\}@hanyang.ac.kr}}

\begin{document}
\maketitle
\begin{abstract}
World-model video encoders such as V-JEPA learn how scenes change---direction, speed, order, state---from video alone.
To answer questions, they are joined to a language model trained on captions and question--answer pairs.
But each caption or question names only a few attributes of a clip, so the model learns to retrieve those attributes rather than to read the clip, and the dynamics the encoder carries go unused: on the same items, a linear probe reads speed from frozen V-JEPA features at 87\% while the VLM answers at chance.
We show where dynamics lives in a frozen encoder: in a subspace shared across videos for predictive encoders such as V-JEPA, but not as a shared component of frame-wise, text-aligned encoders such as SigLIP. We then build counterfactual supervision that makes the language model read it: pixel transformations turn public videos into answer-balanced variants whose answers are set by the transformation and differ only in the change.
Trained in a single stage on 4.53M public samples with 32 visual tokens per frame, where open video LLMs use several stages, 9M to 172M samples and 64 to 256 tokens per frame on average, the resulting \ours models (1B to 8B, Qwen3 and Llama) lead every size class in average score across eleven video benchmarks; our 4B model scores 72.1, against 61.1 for the best open 4B and 64.8 for the best open 8B model.
A controlled encoder\,$\times$\,supervision study shows that the encoder decides what can be read and the supervision what is read: at a matched budget, counterfactual supervision adds 4.5 points on average and 38 on the speed questions of natural benchmarks.
Code and models: \url{https://github.com/kkt94/Llama-V-JEPA-LM}.
\end{abstract}

\section{Introduction}
\label{sec:intro}

\begin{figure*}[t]
\centering
\includegraphics[width=\textwidth]{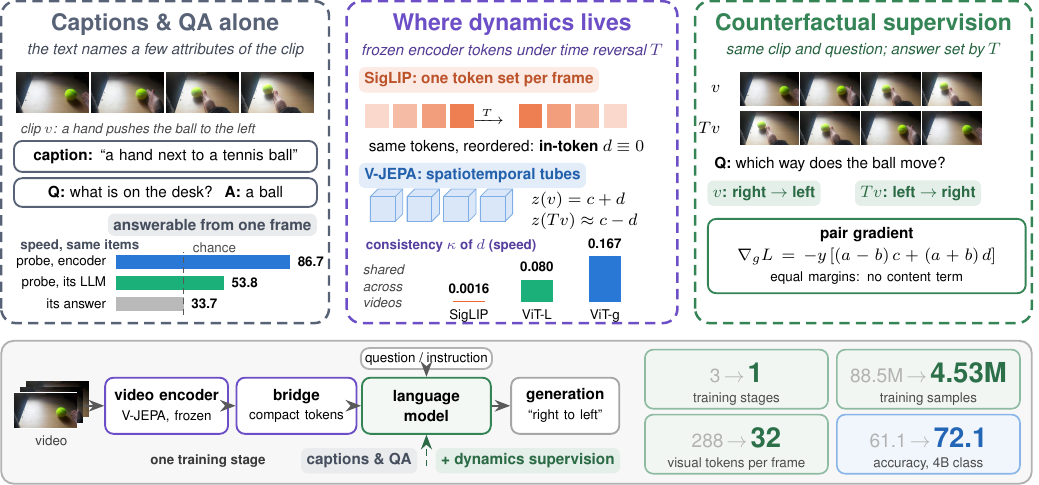}
\caption{Top: what captions and QA teach (left), where the dynamics lives in the encoder (middle), and the supervision built from it (right). Bottom: the model, trained in one stage, and what the recipe changes (before $\to$ after: the V-JEPA~2 video LLM~\cite{assran2025vjepa2}; accuracy: the best open 4B model).}
\label{fig:overview}
\end{figure*}

Self-supervised video encoders trained to predict masked spatiotemporal regions in latent space, the V-JEPA family~\cite{assran2023ijepa,bardes2024vjepa,assran2025vjepa2,murlabadia2026vjepa21}, learn how the physical world moves from video alone, without a single caption: they are world models whose features support motion understanding, anticipation and planning.
To answer in language, however, V-JEPA~2 and 2.1 are turned into video LLMs with the recipe of image--text VLMs~\cite{liu2023llava,qwen3vl,internvl35,cho2025plm}: alignment, pre-training and instruction stages on captions and question--answer pairs, 88.5M and 72.5M samples, and 288 tokens per frame for V-JEPA~2~\cite{assran2025vjepa2,murlabadia2026vjepa21}.
The model built on a world-model encoder is thus aligned to text once more, and only the encoders are released, so how a language model can read what such an encoder knows remains open.

Scale does not make the model read dynamics.
Each caption or question names a few attributes of its clip, and an answer that matches them lowers the loss, so the model learns to fetch what the text asks about rather than to read the clip, and most such items can be answered from one frame~\cite{lei2023singleframe}.
On held-out speed-ratio items, a linear probe reads the answer from frozen V-JEPA features at 87\% and from the bridge of a VLM trained on the caption and QA mixture at 84\%, yet that VLM answers at chance (\cref{sec:unused}).
A stronger encoder does not fix this: in Meta's controlled comparison, V-JEPA~2.1 lowers TemporalBench and TOMATO relative to V-JEPA~2 under the same recipe~\cite{murlabadia2026vjepa21}.
What is missing is supervision that requires the change.

We therefore design the supervision from where the dynamics lives in the encoder (\cref{sec:analysis}).
For a frame-wise, text-aligned encoder such as SigLIP, time reversal merely reorders the tokens, so the arrow of time is in no token's content. In V-JEPA, reversing, mirroring or re-timing a clip moves its tokens along a change component that points the same way across videos, 20 to 100 times more consistently than in SigLIP; a text-aligned video encoder with attention across frames behaves like SigLIP.
When a clip and a transformed copy are asked the same question and their answers differ by construction, only the change can tell them apart, so the content they share cannot separate the answers (\cref{fig:overview}).
Fourteen such constructions in six families are rendered from public videos by pixel transformations, without new annotation; adding a video collection takes a short script that names its clips and the constructions to apply (\cref{sec:method}).

Far less data, better models.
\ours trains in a single stage on 4.53M public samples with 32 visual tokens per frame, where open video LLMs use several stages, 9M to 172M samples and 64 to 256 tokens per frame on average; for every model we count the vision--language training that turns a pretrained vision encoder and a pretrained language model into the VLM (\cref{fig:efficiency}).
With 2 to 38 times less data it leads every size class on eleven video benchmarks, and its 8B models lead Meta's V-JEPA video LLMs with a twentieth of the data. At a matched budget, counterfactual supervision adds 4.5 points to the 2B model's average and 38 on speed questions, while annotated temporal questions or the arrow of time at the same weight do not.

Our contributions are as follows.
\begin{itemize}[leftmargin=*,itemsep=1pt,topsep=2pt]
\item Where dynamics lives: a measured analysis separating frame-wise text-aligned from predictive encoders, which tells what supervision a predictive encoder can use.
\item Counterfactual supervision and an extensible generator: answers fixed by pixel transformations of public videos, so content cancels out; any public video collection becomes training data with a short script and no annotation.
\item Data-efficient video LLMs: \ours, 1B to 8B on Qwen3 and Llama, one stage, 4.53M samples, first in every size class; weights and inference code are released, and training code, data generator and manifests will follow.
\item A controlled encoder\,$\times$\,supervision study with matched-weight controls, held-out generalization and internal measurements.
\end{itemize}

\noindent Supplementary material: proofs and \cref{prop:unpaired,prop:kappa} in \cref{app:theory}; the 33 counterfactual sources, examples, rendering, label audit and closest work in \cref{app:data}; the mixture, settings, curves, protocol, reported against measured scores, train--evaluation audit, held-out sets, successive playback, single-frame control and inference cost in \cref{app:train}; encoder measurements in \cref{app:enc}; every 2$\times$2 checkpoint, pooled differences, ability groups and the AoT, temporal-QA and 4B controls in \cref{app:2x2}; reversal and mirror consistency in \cref{app:heldout}; answer-position measurements and subspace removal in \cref{app:internal}; explicit pairs in \cref{app:pairs}; scope in \cref{app:limits}.

\section{Related Work}
\label{sec:related}

\paragraph{Video LLMs and encoders.}
Open video LLMs~\cite{maaz2024videochatgpt,lin2024videollava,li2025llavaonevision,zhang2025llavavideo,qwen3vl,internvl35,molmo2,cho2025plm,videollama3,gemma4} pair contrastive image encoders~\cite{radford2021clip,zhai2023siglip,tschannen2025siglip2} with a language model and train on large caption and VQA corpora.
Encoder studies favor SigLIP, with video encoders helping mainly on temporal perception~\cite{zohar2025apollo,tong2024cambrian,fan2025webssl}.
V-JEPA~2 and 2.1 attach their encoders to Llama~3.1~8B with this recipe~\cite{assran2025vjepa2,murlabadia2026vjepa21}; VL-JEPA predicts text embeddings rather than tokens~\cite{chen2025vljepa}; we keep the language model as the answer head and change the supervision rather than the scale.

\paragraph{Dynamics present but unread; temporal training signals.}
Motion direction remains linearly decodable inside video LLMs while the answer ignores it, and a projector-level motion objective helps~\cite{lee2026whichway}.
Closest to us, Tracing the Arrow of Time~\cite{han2026tracing} finds arrow-of-time information in video-centric but not frame-centric encoders, shows that query-based connectors disrupt it, and trains with a time-preserving MLP and arrow-of-time supervision (TVBench $+1.3$).
We share the encoder observation and go beyond one binary target: a theory of which transformations give a shared, content-free component, fourteen answer-balanced constructions in six families, and a controlled encoder\,$\times$\,supervision study; an arrow-of-time-only variant of our data is compared in \cref{sec:2x2}.
Reversed, re-timed and shuffled clips have long served as self-supervised pretext tasks for learning video features~\cite{misra2016shuffle,fernando2017odd,wei2018arrow,xu2019clip,benaim2020speednet,yao2020prp,wang2020pace} and as temporal benchmarks and probes~\cite{buch2022revisiting,bagad2023testoftime,li2024vitatecs,liu2024tempcompass,shangguan2025tomato,zhang2024vinoground,krojer2025mvp,cores2024tvbench,cai2024temporalbench,hong2025motionbench}; we apply them to a frozen encoder, as answer-balanced questions that teach a language model to read features the encoder already has.
Training signals include perturbed-clip negatives~\cite{li2025temple}, temporal preferences~\cite{li2025tpo}, frame-order fragments~\cite{hu2025sf2t}, an arrow-of-time reward~\cite{xue2025arrow} and generated counterfactual videos~\cite{poppi2026countervid}, each using one or a few transformations; SynRL~\cite{jiang2026synrl} renders synthetic shapes with exact labels for CoT and RL training. TIME~\cite{wang2026time} adds answer-balanced temporal instructions in five dimensions, with reversed videos for direction. We instead set each answer by a transformation of the same clip, so scene content cancels from it, choose transformations by structure measured in the frozen encoder, and add them to a mixture that already holds annotation-based temporal questions (\cref{sec:2x2}).

\section{Where the Dynamics Lives in a Video Encoder}
\label{sec:analysis}

We derive when an encoder carries a shared change component and measure it, because this decides how efficiently a language model learns to use the dynamics.

\begin{figure}[t]
\centering
\includegraphics[width=\linewidth]{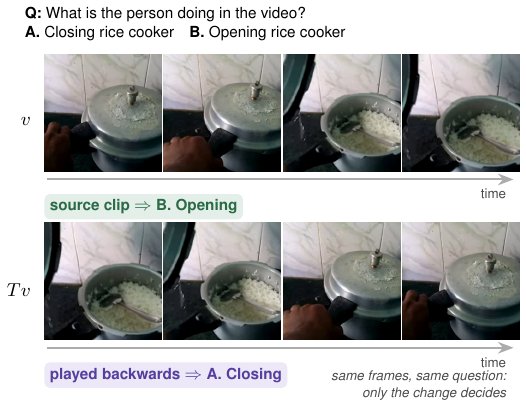}
\caption{One generated pair: the same clip forward and backward, with the same question and opposite answers (4 of 32 frames). Other constructions: \cref{fig:app_examples}.}
\label{fig:example}
\end{figure}
\paragraph{Setup.}
Let $z(v)$ be the frozen encoder tokens of a video $v$, and $T$ a pixel transformation with an answer map $\pi_T$ and token permutation $\Pi_T$, so that $y(Tv)=\pi_T(y(v))$.
With $z_T(v)=\Pi_T^{-1}z(Tv)$, the tokens of $Tv$ at the positions of $v$, we define the content $c=\frac12(z(v)+z_T(v))$ and the change $d=\frac12(z(v)-z_T(v))$.
We describe $d$ by its relative size $r_d=\mathbb{E}\lVert d\rVert^2/\mathbb{E}\lVert z-\bar z\rVert^2$, its overlap $\rho$ with the leading content directions, and its consistency across videos $\kappa=\lVert\mathbb{E}[y\,d]\rVert^2/\mathbb{E}\lVert d\rVert^2$, with $y=\pm1$ the direction of the change: 1 if every video moves by the same vector, about $0.001$ for random directions.

\begin{proposition}[frame-wise, text-aligned encoders]
\label{thm:b}
(i) If tokens are functions of single frames, $E(v)=[f(x_1),\dots,f(x_N)]$, time reversal only reorders them, so the in-token change is $d\equiv0$ exactly: the arrow of time is not in any token's content and can only be computed from token positions; a speed change selects other frames, so $d$ is a scene-specific frame difference.
(ii) If the caption distribution is invariant to mirroring, a contrastive loss is too, so for an unnormalized (convex) loss a minimizer can be mirror-invariant in the caption-aligned subspace; for normalized embeddings we measure it ($\kappa$, \cref{tab:enc}).
\end{proposition}

\begin{proposition}[predictive spatiotemporal encoders]
\label{thm:a}
Assume the training loss is invariant under $T$ (data, masks and the predictor's positional coding are symmetric) and that minimizers differ only by an invertible linear map on token channels, the same at every position (identifiability~\cite{roeder2021linear}, assumed here). Then $\Pi_T^{-1}E(Tv)=A_TE(v)$ with $A_T$ acting on channels and $T\mapsto A_T$ a group representation.
For an involution the features split into fixed subspaces $V_+\oplus V_-$ with $c\in V_+$ and $d\in V_-$ for every video. If a linear readout $w$ that weights all token positions alike ($\Pi_T^\top w=w$) answers a question whose answer flips under $T$, its part $w_-=\frac12(w-A_T^\top w)$ gives the same answer on every video and ignores content ($w_-\cdot c=0$). Here ``content'' means the $T$-invariant component, not the scene's semantics.
\end{proposition}

\noindent Both concern token content, not relations between tokens, the route left to a frame-wise encoder; we measure their consequences, $\kappa$ and $\rho$, directly (order, speed and proofs: \cref{app:theory}).

\begin{table*}[t]
\centering\footnotesize\setlength{\tabcolsep}{3.2pt}
\begin{tabular}{>{\raggedright\arraybackslash}p{1.35cm}>{\raggedright\arraybackslash}p{5.4cm}>{\raggedright\arraybackslash}p{4.1cm}>{\raggedright\arraybackslash}p{3.85cm}rr}
\toprule
Family & Construction of $Tv$ & Answer rule $\pi_T$ & Source videos & Rows & Pair\% \\
\midrule
Sign flip & State change played backwards & action $\leftrightarrow$ inverse (open/close) & Charades-Ego, EPIC, SSv2 & 49k & 3 \\
 & Object direction, clip mirrored or reversed & direction remapped & SSv2 & 37k & 0 \\
 & Camera pan, tilt or zoom on static footage, or reversed & move remapped & CLEVRER, Charades-Ego, SSv2 & 23k & 0 \\
 & Two clips joined, each optionally reversed & direction A then B & SSv2 & 7k & 0 \\
\midrule
Order & Two events in sequence or stacked at once & first / second / simultaneous & LLaVA-Video, SA-V, SSv2 & 65k & 0 \\
 & Segment matching a two-action caption & matching segment & HoloAssist & 20k & 80 \\
\midrule
Position & Joined clips; one holds the queried action & beginning / middle / end & HoloAssist, EPIC, Kinetics-700 & 48k & 68 \\
\midrule
Magnitude & One clip stacked twice over unequal spans & faster / same / slower & SA-V, SSv2 & 81k & 0 \\
 & Span split 1:3, 3:1 or 1:1 between halves & accelerates / decelerates / steady & SA-V, SSv2 & 48k & 0 \\
 & Frames spread over a short or long window & slow motion / normal / time-lapse & EPIC & 12k & 0 \\
\midrule
Count & One action segment repeated $k$ times & $k$ & HoloAssist, EPIC & 24k & 86 \\
 & Segments of 1--4 different activities & number of activities & Kinetics-700 & 12k & 100 \\
\midrule
Stop/cut & Clip cut before the action completes & complete / incomplete & Charades-Ego & 16k & 5 \\
 & Simulator physics edits (teleport, vanish) & possible / impossible & simulator & 66k & 100 \\
\bottomrule
\end{tabular}
\caption{The fourteen counterfactual constructions (33 sources, 509k rows, 6.1\% of sampled training rows). Each row is rendered at training time from a clip of a public dataset and a transformation; the answer follows from $\pi_T$. Pair\%: rows whose counterpart from the same clip with a different answer is also in the same source. Per-source details, real questions and rendered examples: \cref{app:data}.}
\label{tab:constructions}
\end{table*}

\begin{table}[t]
\centering\footnotesize\setlength{\tabcolsep}{2.6pt}
\begin{tabular}{lrrrrrr}
\toprule
& \multicolumn{2}{c}{reversal} & speed & mirror & \multicolumn{2}{c}{speed readout} \\
\cmidrule(lr){2-3}\cmidrule(lr){4-4}\cmidrule(lr){5-5}\cmidrule(lr){6-7}
Encoder & $r_d$ & $\kappa$ & $\kappa$ & $\kappa$ & proto 8 & probe \\
\midrule
SigLIP so400m & \textbf{0} & -- & .0016 & .0014 & 50.0 & 73.5 \\
ViCLIP (text-aligned) & .052 & .0012 & .042 & .0019 & 60.1 & 70.3 \\
VideoMAE (pixel pred.) & .502 & .0032 & .632 & .045 & 100.0 & 76.8 \\
\midrule
\rowcolor{ours}V-JEPA ViT-L (1B, 2B) & .064 & .037 & .080 & .031 & 66.3 & 81.2 \\
\rowcolor{ours}V-JEPA ViT-g (3B--8B) & .108 & .039 & .167 & .074 & 77.2 & 84.7 \\
\bottomrule
\end{tabular}
\caption{Change component of frozen encoders (1,200 clips; 70\% fit, 30\% held out). $r_d$: relative size of the in-token change after aligning token positions; $\kappa$: cross-video consistency; proto 8: accuracy of a prototype read from 8 pairs (chance 50); probe: linear probe on all training clips. ViCLIP and VideoMAE: diagnostic encoders with joint space--time attention, not used in our models. All measures: \cref{tab:app_enc}.}
\label{tab:enc}
\end{table}

\begin{table}[t]
\centering\footnotesize\setlength{\tabcolsep}{3.2pt}
\begin{tabular}{@{}llrrrrrr|r@{}}
\toprule
 & & \multicolumn{6}{c|}{linear probe on the stage} & VLM \\
Encoder & Training & frozen & bridge & L4 & L14 & L18 & L28 & answer \\
\midrule
SigLIP & caps+QA & 36.4 & 38.7 & 36.8 & 35.9 & 35.7 & 34.7 & 33.1 \\
SigLIP & +CF & 36.3 & 39.1 & 37.5 & 37.1 & 35.9 & 35.8 & 34.4 \\
\midrule
V-JEPA & caps+QA & 86.7 & 83.5 & 53.8 & 47.2 & 49.4 & 46.9 & 33.7 \\
\rowcolor{ours}V-JEPA & +CF & 86.8 & 91.2 & 66.6 & 78.9 & 96.3 & 96.5 & 92.8 \\
\bottomrule
\end{tabular}
\caption{One item set at every stage: the 540 held-out speed-ratio items (unseen ratios 1.6 / 2.5 / 4.0 and equal; SSv2 validation; 3 classes, chance 33.3), 2B models at 8k steps. Probe: linear (PCA 64, logistic) at the answer position for language-model layers and on a $16{\times}2{\times}2$ token grid for the frozen encoder and the bridge, 5-fold cross-validation grouped by source clip, 4 probe seeds (s.d.\ ${\le}1.8$). Answer: the VLM on the same inputs (one option order per clip), within 2 points of the evaluation in \cref{tab:2x2}.}
\label{tab:unused}
\end{table}

\subsection{Testing the two propositions on video encoders}
\label{sec:measure}
We test them on the frozen features of the encoders of our study, SigLIP so400m and V-JEPA~2.1 ViT-L and ViT-g, and of two more video encoders with joint space--time attention, ViCLIP~\cite{wang2024internvid} (text-aligned) and VideoMAE~\cite{tong2022videomae} (pixel prediction).
We use 1,200 SSv2 training clips that appear in no evaluation set, decoded as in training, each paired with its reversal, a half-speed version and its mirror image, so $d$ holds only what token content says about $T$.
Four measures ask whether a change component exists ($r_d$), whether it is shared across clips ($\kappa$), whether 8 pairs read it on held-out clips (proto~8) and whether it is present at all (probe).
In \cref{tab:enc}, SigLIP has $r_d=0$ under reversal exactly, and its speed and mirror components have $\kappa<0.002$, so 8 pairs give chance; a probe that sees token positions still reads time direction (65.3\%) and speed (73.5\%), the relational route of \cref{thm:b}.
Both V-JEPA encoders carry a change component with $\kappa$ 20 to 100 times larger, readable from 8 pairs, and the larger encoder is better on every measure.
Reversal and mirror changes are nearly orthogonal to content ($\rho$ 0.03--0.12, against 0.44--0.63 for a content direction); the speed change is the most consistent but overlaps content ($\rho$ 0.40 / 0.59), which answer balance must separate (\cref{sec:method}).
Attention across frames does not by itself give this structure: text-aligned ViCLIP has no shared reversal or mirror direction either ($\kappa\le.002$), pixel-predicting VideoMAE shares speed and mirror changes but not reversal ($\kappa=.003$), and only V-JEPA shares all three and reads time direction and speed best (probes 80.8 and 81.2--84.7; \cref{tab:app_enc}).
V-JEPA's sign-flip and speed changes thus need only a few counterfactual pairs to fix their direction (\cref{prop:kappa}), whereas SigLIP must learn them from token relations; order moves token positions in both encoders and should help both alike (\cref{tab:app_map}).

\subsection{Present in the encoder, unused by the VLM}
\label{sec:unused}
Does a VLM built on this encoder with the usual caption and QA mixture use the change?
\Cref{tab:unused} follows one item set through every stage: the 540 held-out speed-ratio items of \cref{sec:gen} (unseen ratios, three classes, chance 33.3), read by a linear probe on the frozen encoder, on the bridge output and on language-model layers at the answer position, and answered by the VLM itself.
For V-JEPA, the frozen features give 86.7\% and the bridge of the caption and QA model keeps 83.5\%, yet at the answer position 47--54\% remains and the answer is at chance (33.7\%); the model gives the same answer to a clip and its re-timed counterpart 89\% of the time.
With the counterfactual rows of \cref{sec:method}, the same encoder and mixture give 91.2\% at the bridge, 96\% from layer 18 on, and a 92.8\% answer.
For SigLIP no stage exceeds 40\%, with or without the rows, as \cref{thm:b} and $\kappa\approx0.002$ predict.
The information reaches the language model, but captions and questions do not require it, and the answer does not use it. Nor is the item format the obstacle: SigLIP trained on the same counterfactual rows reaches only 34.8 on these items at 8k steps and 55.7 at the end, against 98.7 for V-JEPA (\cref{sec:gen}), and on TOMATO's natural speed questions, posed in another format, the rows add 62.9 points (\cref{sec:skills}).

\section{Counterfactual Supervision}
\label{sec:method}

\Cref{sec:unused} shows the change reaching the language model without being required. The caption and QA mixture stays as it is and keeps teaching language and knowledge; we add rows that require the change.

\paragraph{Constructions with answers set by the transformation.}
A generator takes a clip $v$ from a public dataset and a transformation $T$, renders $Tv$, and writes the question, options and answer $\pi_T(y)$, which follows from the transformation: no new manual annotation and no language model are involved (constructions on actions and events reuse the source datasets' own labels).
We use fourteen constructions in six families (\cref{tab:constructions}): sign flips (reversal, mirroring, camera moves), event order, temporal position, speed and rate, counts, and stopped or impossible events.
Within each source the variants of a clip are drawn so that every scene appears with each answer equally often in expectation; for 29\% of the rows a counterpart from the same clip, source and question family with a different answer is also in the data, an explicit pair (\cref{fig:example,app:pairs}). Rows are sampled independently, so counterparts are not forced into one batch; the speed sources have no explicit pairs and rely on answer balance alone.
The generator also balances the correct option's length rank and each option's prior under the text-only language model, and decorrelates answers from clip length and timing.

\begin{table*}[t]
\centering\footnotesize\setlength{\tabcolsep}{3.6pt}
\begin{tabular}{lrrrrrrrrrrrrr}
\toprule
Model & MVB & PT & TVB & TC & TOM & MoB & NQA & MVP & TB & SSv2 & VG-T & VG-V & Avg \\
\midrule
\multicolumn{14}{l}{\textit{2B class}} \\
Qwen3-VL-2B & 61.0 & 65.1 & 49.8 & 67.4 & 30.3 & \textbf{57.5} & 77.1 & 32.3 & 65.9 & 59.2 & 36.2 & 27.6 & 52.45 \\
InternVL3.5-2B & 66.8 & 64.2 & 50.9 & 65.7 & 31.0 & 56.2 & 76.1 & 28.8 & \textbf{71.3} & 57.0 & 40.0 & 21.0 & 52.42 \\
PLM-1B & 68.0 & 69.8 & 48.2 & 60.5 & 24.5 & 51.7 & \textbf{79.8} & 30.9 & 58.1 & 61.7 & 33.2 & 29.6 & 51.33 \\
Apollo-1.5B & 61.1 & 63.9 & 47.9 & 60.3 & 27.8 & 51.1 & 78.8 & 26.0 & 55.8 & 56.0 & 38.8 & 22.4 & 49.16 \\
VideoLLaMA3-2B & 57.9 & 62.9 & 44.0 & 64.1 & 24.0 & 56.3 & 75.4 & 23.1 & 64.1 & 50.0 & 38.0 & 25.2 & 48.75 \\
\rowcolor{ours}Ours-2B-Q (Qwen3-1.7B) & \textbf{69.5} & \textbf{75.3} & \textbf{70.7} & \textbf{73.5} & \textbf{50.9} & 54.3 & 74.0 & \textbf{54.6} & 63.9 & \textbf{91.6} & \textbf{63.2} & 50.8 & \textbf{66.02} \\
\rowcolor{ours}Ours-1B-L (Llama-3.2-1B) & 66.3 & 71.9 & 67.9 & 71.6 & 48.2 & 49.5 & 70.2 & 48.5 & 57.8 & 89.6 & 51.2 & \textbf{53.8} & 62.21 \\
\midrule
\multicolumn{14}{l}{\textit{4B class}} \\
Molmo2-4B & \textbf{74.2} & \textbf{81.4} & 65.5 & 70.8 & 36.5 & \textbf{62.3} & \textbf{85.1} & 45.4 & 72.8 & 57.0 & 48.2 & 33.8 & 61.08 \\
PLM-3B & 73.0 & 76.0 & 56.6 & 67.0 & 36.3 & 58.9 & 84.0 & 46.2 & 66.2 & 68.3 & 55.0 & 31.4 & 59.91 \\
InternVL3.5-4B & 70.9 & 69.1 & 60.0 & 71.1 & 32.1 & 59.0 & 79.8 & 41.4 & 73.2 & 64.1 & 53.6 & 28.0 & 58.52 \\
Qwen3-VL-4B & 68.0 & 72.0 & 59.2 & 70.9 & 36.3 & 61.5 & 80.9 & 39.6 & 68.9 & 64.8 & 47.6 & 31.0 & 58.39 \\
Apollo-3B & 62.0 & 67.1 & 50.9 & 63.0 & 29.0 & 55.0 & 80.9 & 27.4 & 60.1 & 64.7 & 37.4 & 25.8 & 51.94 \\
\rowcolor{ours}Ours-4B-Q (Qwen3-4B) & 72.1 & 78.2 & \textbf{74.8} & \textbf{79.9} & 53.5 & 59.1 & 79.3 & \textbf{58.1} & \textbf{75.7} & \textbf{92.2} & \textbf{76.0} & \textbf{66.8} & \textbf{72.14} \\
\rowcolor{ours}Ours-3B-L (Llama-3.2-3B) & 70.7 & 74.3 & 70.3 & 76.5 & \textbf{55.9} & 54.2 & 77.0 & 56.3 & 67.3 & 91.3 & 71.0 & 62.8 & 68.97 \\
\midrule
\multicolumn{14}{l}{\textit{8B class}} \\
PLM-8B & \textbf{76.0} & 80.6 & 62.7 & 70.9 & 42.6 & 62.7 & \textbf{85.7} & 49.0 & 71.2 & 67.9 & 63.4 & 44.8 & 64.79 \\
Molmo2-8B & 75.6 & \textbf{81.9} & 64.1 & 72.1 & 35.4 & 61.4 & 85.4 & 43.4 & 70.8 & 55.2 & 49.2 & 33.8 & 60.69 \\
Qwen3-VL-8B & 68.8 & 73.0 & 58.2 & 73.6 & 36.5 & \textbf{62.9} & 82.1 & 41.8 & 70.8 & 65.9 & 51.0 & 35.0 & 59.97 \\
InternVL3.5-8B & 71.0 & 72.7 & 62.7 & 71.0 & 33.6 & 58.1 & 81.5 & 38.8 & 72.1 & 66.7 & 47.2 & 28.4 & 58.65 \\
VideoLLaMA3-7B & 65.6 & 69.7 & 49.2 & 69.3 & 29.4 & 57.8 & 81.5 & 29.9 & 70.5 & 59.1 & 44.8 & 28.6 & 54.62 \\
\rowcolor{ours}Ours-8B-Q (Qwen3-8B) & 72.8 & 77.8 & \textbf{75.3} & \textbf{79.4} & \textbf{56.5} & 59.1 & 81.2 & \textbf{58.1} & \textbf{73.5} & \textbf{93.8} & \textbf{75.2} & 38.6 & 70.11 \\
\rowcolor{ours}Ours-8B-L (Llama-3.1-8B) & 71.9 & 75.0 & 71.8 & 77.6 & 55.5 & 55.9 & 79.0 & 57.6 & 72.3 & 92.3 & 74.2 & \textbf{63.0} & \textbf{70.51} \\
\bottomrule
\end{tabular}
\caption{Main results: twelve scores on eleven video benchmarks. All models evaluated by us with one protocol (32 frames, same prompts, option-letter scoring). Shaded: \ours (-Q: Qwen3, -L: Llama). Bold: best in class. MVB: MVBench, PT: PerceptionTest, TVB: TVBench, TC: TempCompass, TOM: TOMATO, MoB: MotionBench, NQA: NExT-QA, MVP and VG-T/VG-V (Vinoground text/video): pair accuracy, TB: TemporalBench (binary), SSv2: Something-Something v2 (multiple choice). 95\% bootstrap intervals of the average: $\pm$0.7 for every model (\cref{app:train}). Gemma~4, last in both classes: \cref{tab:app_gemma}.}
\label{tab:main}
\end{table*}

\paragraph{Why the construction makes the model read the change.}
Let $s(v)$ be the answer log-odds at the scoring position and $g=\partial s/\partial z$ the readout through bridge and language model; in the linear regime a pair splits into a common part $m=g\cdot c$ and a differential part $d_s=g\cdot d$.

\begin{proposition}[pair loss and content]
\label{prop:pair}
For $L=\ell(y\,s(v))+\ell(-y\,s(Tv))$ with logistic $\ell$, (i) $\nabla_g L=-y[(a-b)\,c+(a+b)\,d]$ with $a=-\ell'(y\,s(v))$, $b=-\ell'(-y\,s(Tv))$, so at equal margins the content coefficient is exactly zero; (ii) for fixed $d_s$, $L$ is even and convex in $m$, hence minimized at $m=0$.
If the readout can set $g\cdot c$ and $g\cdot d$ independently (content and change separately addressable, \eg $c\in V_+$, $d\in V_-$ as in \cref{thm:a}), the optimum of the pair loss therefore has no content in the answer: $g\cdot c=0$.
\end{proposition}

\noindent Without separability $m=0$ is not implied (\cref{app:theory} gives an example), so we measure the content share at the answer (\cref{sec:inside}). For binary answers it extends to any answer-balanced set at equal error weights, in expectation for separate rows and within each scene for explicit pairs.

Without pairs, a scene--answer correlation gives the readout a content component (\cref{prop:unpaired}); from a zero readout the first step on $k$ pairs is the prototype $\sum_i y_i d_i$, which reads a new pair's change better as $\kappa$ grows and a single clip better as $\rho$ shrinks (\cref{prop:kappa}; both in \cref{app:theory}).

\noindent With the measured $\kappa$, this anticipates large gains from sign-flip and speed constructions on V-JEPA, small ones on SigLIP and equal gains from order on both (\cref{tab:app_map}), and a change in how the answer layers are organized more than in what a probe can read at the bridge input (\cref{sec:inside}).

\paragraph{Model and training.}
A frozen V-JEPA~2.1 encoder (ViT-L for 1B and 2B, ViT-g for 3B--8B) reads 32 frames; a spectral bridge maps each clip to 1,024 tokens: a fixed PCA rotation and rescaling computed once from frozen encoder statistics, followed by a trainable MLP projector.
After a bridge-only warm-up of 0.6--0.8\% of steps, the language model~\cite{qwen3,llama3} is trained in full in a single stage~\cite{karamcheti2024prismatic}, in place of the usual alignment, pre-training and instruction stages, on 4.53M rows drawn at fixed per-source weights from the training splits of 175 public sources, the caption and QA collections~\cite{zhang2025llavavideo,zhang2024llavahound,cho2025plm} plus the 33 counterfactual sources (about 6\% of drawn rows), at global batch 150--300 (settings: \cref{tab:app_train}; curves and metrics: \cref{fig:app_train_curves,tab:app_train_stats}).

\section{Experiments}
\label{sec:exp}

\paragraph{Benchmarks and protocol.}
We start from the benchmarks of Meta's V-JEPA video LLMs~\cite{assran2025vjepa2,murlabadia2026vjepa21}, MVBench~\cite{li2024mvbench}, PerceptionTest~\cite{patraucean2023perception}, TVBench~\cite{cores2024tvbench}, TempCompass~\cite{liu2024tempcompass} (multiple choice), TOMATO~\cite{shangguan2025tomato}, MVP~\cite{krojer2025mvp} (pair accuracy) and TemporalBench~\cite{cai2024temporalbench} (binary), and extend them to twelve scores from eleven benchmarks with MotionBench~\cite{hong2025motionbench}, NExT-QA~\cite{xiao2021nextqa}, SSv2~\cite{goyal2017ssv2} (multiple choice over action labels), and Vinoground~\cite{zhang2024vinoground} text and video (pair accuracy).
The additions cover motion (MotionBench), causal and temporal reasoning (NExT-QA), fine-grained actions (SSv2) and temporal order that a single frame cannot answer (Vinoground).
We evaluate every open model ourselves with the same protocol: 32 frames, the same prompts and option-letter scoring; their own reported numbers, under their settings, are in \cref{tab:app_reported}. We report the plain average of the twelve.

\begin{table}[t]
\centering\footnotesize\setlength{\tabcolsep}{2.6pt}
\resizebox{\linewidth}{!}{\begin{tabular}{lrrrrrrrrr}
\toprule
Model & Tok. & Samples & MVB & PT & TVB & TC & TOM & MVP & Avg \\
\midrule
V-JEPA 2 (L3.1-8B) & 9,216 & 88.5M & 73.5 & 84.0 & 60.6 & 76.9 & 40.3 & 44.5 & 63.3 \\
V-JEPA 2.1 (L3.1-8B) & -- & 72.5M & 75.4 & 83.1 & 62.8 & 74.0 & 38.0 & 43.2 & 62.8 \\
\rowcolor{ours}Ours-8B-Q (Qwen3-8B) & 1,024 & 4.53M & 72.8 & 77.8 & 75.3 & 79.4 & 56.5 & 58.1 & 70.0 \\
\rowcolor{ours}Ours-8B-L (L3.1-8B) & 1,024 & 4.53M & 71.9 & 75.0 & 71.8 & 77.6 & 55.5 & 57.6 & 68.2 \\
\bottomrule
\end{tabular}}
\caption{Against Meta's V-JEPA video LLMs on the six benchmarks both report~\cite{assran2025vjepa2,murlabadia2026vjepa21} (L3.1-8B: Llama-3.1-8B, also the language model of Ours-8B-L). Meta values quoted from the papers (the video LLMs are not released, only the encoders): V-JEPA~2 with ViT-g$_{384}$, V-JEPA~2.1 with ViT-G$_{384}$; PerceptionTest is the test split after training on its train split for V-JEPA~2 and the validation split for V-JEPA~2.1. Ours measured.}
\label{tab:meta}
\end{table}

\paragraph{Main results.}
\Cref{tab:main} lists every benchmark. With one training stage, a fraction of the data and 32 tokens per frame, our models have the highest average in each size class: the 4B model (72.14) is 11.1 points above the best open 4B model and 7.4 above the best open 8B model, the 2B model (66.02) is 13.6 above the best open 2B model, and the Llama models lead alike (68.97 on Llama-3.2-3B and 62.21 on Llama-3.2-1B). Without SSv2 and the two Vinoground scores every model still leads its class (4B: 70.1 against 66.0).
The largest margins are on the dynamics benchmarks (4B class: TOMATO 53.5 against at most 36.5, Vinoground video 66.8 against at most 33.8), and our 4B model answers both the original and the reversed or mirrored version of natural items correctly on 47.7\% and 50.4\%, against at most 35.4\% for open 4B models (\cref{tab:app_cf}).
Against Meta's unreleased video LLMs on the same encoder family (numbers quoted from their papers; \cref{tab:meta}), our 8B models lead on their six benchmarks by 6.7 (Qwen3-8B) and 4.9 (Llama-3.1-8B, Meta's own).

\begin{figure}[t]
\centering
\includegraphics{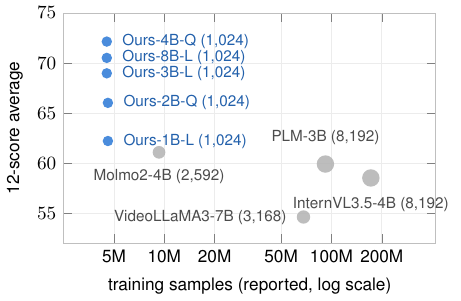}
\caption{Twelve-score average against training samples as reported (log scale). Marker area: visual tokens per clip. Sample counts: \cref{app:train}.}
\label{fig:efficiency}
\end{figure}

\paragraph{Data and compute.}
For every model we count the samples of the vision--language training that turns a pretrained vision encoder and a pretrained language model into the VLM, all of its stages as each paper reports them (for Molmo2 only its SFT, a lower bound); the pre-training of the vision encoder (V-JEPA~2.1 here, SigLIP or InternViT in the open models) and of the language model is excluded alike (\cref{app:train}). On this count our single stage uses 4.53M public samples (4.60M for ViT-L), 2 to 38 times fewer than the open models that report counts (\cref{fig:efficiency}), with 2 to 8 times fewer visual tokens, in 499 H100-hours (4B) or 229 (2B).
At inference (\cref{tab:app_speed}; H100) the bridge reduces the encoder's 9,216 patches to 1,024 tokens, so the language model prefills with 2 to 18 times fewer FLOPs than open models of the same size, more than our whole encoder (14 or 38 TFLOPs) against models with 8,192 tokens; the first token takes 70--182\,ms at any resolution. Each model is first in its class at every matched sample count (\cref{tab:sizes}).

\subsection{Encoder $\times$ supervision}
\label{sec:2x2}
\begin{figure}[t]
\centering
\includegraphics{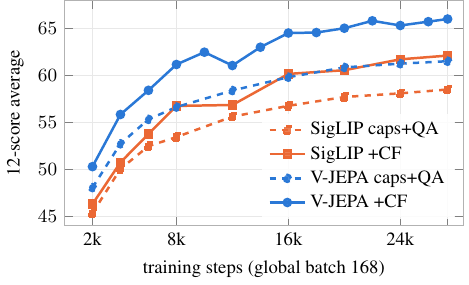}
\caption{Twelve-score average during training (2B). Solid: with counterfactual rows; dashed: captions and QA only.}
\label{fig:traj}
\end{figure}

Which part of the gain is the encoder's and which the supervision's? Four 2B models cross V-JEPA ViT-L or SigLIP so400m with or without the counterfactual rows, on the same mixture and schedule (\cref{fig:traj,tab:2x2}).

\begin{table*}[t]
\centering\footnotesize\setlength{\tabcolsep}{4.2pt}
\begin{tabular}{llrrrrrrrrr}
\toprule
& & \multicolumn{2}{c}{12-score avg} & \multicolumn{4}{c}{abilities at 8k (natural benchmarks)} & \multicolumn{2}{c}{held-out sets at 8k} & \\
\cmidrule(lr){3-4}\cmidrule(lr){5-8}\cmidrule(lr){9-10}
Encoder & Supervision & 8k & final & speed & direction & camera & knowledge & 6-set avg & unseen ratios & rev.\ cons. \\
\midrule
SigLIP so400m & caps+QA & 53.46 & 58.52 & 34.0 & 39.9 & 36.4 & 63.5 & 47.1 & 35.0 & 19.4 \\
SigLIP so400m & +CF & 56.77 & 62.13 & 45.4 & 40.0 & 36.4 & 62.6 & 59.3 & 34.8 & 19.7 \\
V-JEPA ViT-L & caps+QA & 56.65 & 61.53 & 32.4 & 59.7 & 40.5 & 64.1 & 58.6 & 33.5 & 32.3 \\
V-JEPA ViT-L & AoT only & 56.37 & 61.17 & 33.0 & 55.7 & 43.9 & 63.6 & 60.8 & 31.9 & 30.8 \\
\rowcolor{ours}V-JEPA ViT-L & +CF & 61.18 & 66.02 & 65.3 & 63.3 & 47.4 & 63.7 & 91.3 & 91.5 & 36.8 \\
\bottomrule
\end{tabular}
\caption{Encoder $\times$ supervision (2B; same mixture and schedule). +CF: with the counterfactual rows; AoT only: an arrow-of-time question in their place. Final: step 27,382. Abilities: \cref{sec:skills}; held-out chance 33.3; rev.\ cons.: original and reversed video both correct (\cref{tab:app_cf}). All steps and benchmarks: \cref{tab:app_2x2}.}
\label{tab:2x2}
\end{table*}

CF helps both encoders at 8k (4.5 and 3.3 points) and at the end (4.5 and 3.6), and the 4B model by 4.8 at 8k (\cref{app:2x2}); V-JEPA with CF reaches the final SigLIP+CF score (62.13) by 10k steps; the abilities it unlocks differ (\cref{sec:skills}). Against the best open 2B model, all three parts of the recipe add: our mixture and bridge with SigLIP give $+6.1$ (58.52 against 52.45; $+3.2$ without SSv2 and Vinoground), the V-JEPA encoder $+3.0$ and the counterfactual rows $+4.5$.
Matched-weight controls fall short. The arrow-of-time question of \cite{han2026tracing} is learned (95.9\% on unseen clips) but gives 56.37 at 8k and 61.17 at the end, no better than captions and QA (56.65 and 61.53); the mixture's 22 annotation-based temporal-QA sources at that weight give 58.21. Speed stays at the caps-only level (33.0 and 31.3 against 32.4) and the held-out sets at 60.8 and 62.1 (\cref{app:2x2}): the gain comes from the counterfactual construction, not from more temporal questions.

\subsection{Generalization beyond the training constructions}
\label{sec:gen}
\begin{table}[t]
\centering\footnotesize\setlength{\tabcolsep}{2.1pt}
\begin{tabular}{@{}lrrrrrr|r|r@{}}
\toprule
 & new & \multicolumn{2}{c}{new corpus} & \multicolumn{2}{c}{new comb.} & new & & succ. \\
Model & ratio & spd & dir & dir & spd & cam. & avg & play \\
\midrule
Qwen3-VL-2B & 37.0 & 45.5 & 35.2 & 56.5 & 44.6 & 43.9 & 43.8 & 33.1 \\
Qwen3-VL-4B & 35.9 & 50.5 & 40.7 & 73.3 & 50.2 & 49.4 & 50.0 & 36.9 \\
Qwen3-VL-8B & 36.7 & 50.8 & 41.9 & 78.4 & 68.0 & 57.1 & 55.5 & 51.9 \\
V-JEPA, no CF & 33.3 & 44.3 & 69.8 & 97.1 & 49.1 & 92.4 & 64.3 & 39.4 \\
\midrule
\rowcolor{ours}Ours-2B-Q & 98.7 & 98.5 & 75.7 & 98.5 & 98.9 & 99.8 & 95.0 & 93.9 \\
\rowcolor{ours}Ours-4B-Q & 98.9 & 99.0 & 80.9 & 99.0 & 100.0 & 100.0 & 96.3 & 96.7 \\
\rowcolor{ours}Ours-8B-L & 99.3 & 98.7 & 80.0 & 98.8 & 99.6 & 100.0 & 96.1 & 99.2 \\
\bottomrule
\end{tabular}
\caption{Generalization (chance 33.3): the six held-out sets of \cref{sec:gen} and successive playback (succ.\ play). V-JEPA, no CF: the 2B captions and QA arm of \cref{tab:2x2}; like the open models, it never saw these constructions. All models: \cref{tab:app_heldout,tab:app_seq}.}
\label{tab:gen}
\end{table}

Six held-out sets (3,294 items; \cref{tab:gen,tab:app_heldout_def}) each change one factor: speed ratios absent from training (1.6, 2.5, 4.0; training uses equal, 2:1 and 3:1), Kinetics-700, a corpus never used for these families, unseen combinations, and an unseen simulator. Clips come from validation splits or a simulator unused for camera rows, and wordings and option pools are the training generator's own, so only the tested factor is new.
Our models average 94.5--96.3 and open models 43.5--55.5 (\cref{tab:gen}); ratios between and beyond the trained ones are solved alike (97.2, 100 and 99.4 for 1.6, 2.5 and 4.0 on the 4B model; \cref{tab:app_ratio}).
Successive playback changes the construction itself: the clip is played twice in a row, a short central span and the full span, so the motion speeds up or slows down halfway, which no training row uses for speed; our models score 92.5--99.2 against 19.4--51.9 for open models.
In the controlled study V-JEPA with CF reaches 91.3 at 8k steps and 95.0 at the end, against 64.3 without CF and 74.8 for SigLIP with the same rows, which reaches only 55.7 on unseen ratios against 98.7 (\cref{tab:app_heldout}); at 4B, CF likewise lifts the held-out sets from 61.4 to 93.0 and successive playback from 33.1 to 90.0 at 8k steps (\cref{app:2x2}).
Without motion (middle frame repeated) our 2B and 4B models fall to 37.9 and 37.6, so single frames do not solve the sets; apparent motion follows the speed labels on 95--100\% of rows (\cref{app:data}).

\section{Analysis: Why the Supervision Works}
\label{sec:ana}

\subsection{Abilities from the encoder and from the supervision}
\label{sec:skills}
\begin{figure}[t]
\centering
\includegraphics{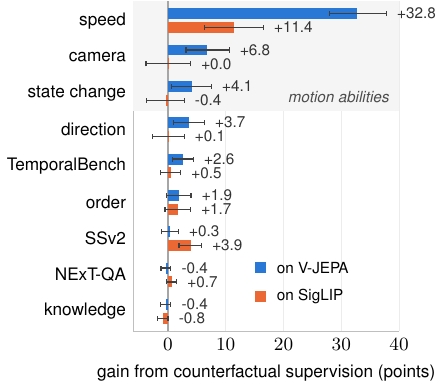}
\caption{Gain from counterfactual supervision per ability (2B, 8k steps); whiskers: 95\% paired bootstrap interval.}
\label{fig:skills}
\end{figure}

The encoder and the supervision contribute different abilities. We group the natural benchmarks' own question categories into abilities (speed, direction, camera motion, state change, \dots) and compare models item by item with a paired bootstrap (\cref{fig:skills}; all groups in \cref{tab:app_skills}; pooled over checkpoints in \cref{tab:app_pool}).
Speed needs both at this budget: without CF the two encoders are level on speed ($-1.5$, interval includes zero); CF adds 32.8 points on V-JEPA (29.4 on the 4B model) and 11.4 on SigLIP (18.0 by 24k); at the end the V-JEPA gain holds on both source benchmarks (TOMATO $+62.9$, TempCompass $+21.8$; \cref{app:2x2}).
Direction, counting and fine actions come from the encoder: without CF SigLIP is already 19.7, 7.7 and 11.5 points behind, and CF barely moves direction and counting on SigLIP.
Camera and state change need a consistent change direction: CF adds 6.8 and 4.1 on V-JEPA and nothing on SigLIP.
General knowledge is kept across encoders ($+0.8$ at 24k, interval includes zero).

\subsection{What the language model reads}
\label{sec:inside}
We test the propositions at the answer position on same-question pairs (\cref{tab:app_internal}).
\Cref{prop:pair}: with CF, 0.94 to 0.98 of the relative-speed answer is differential; without CF the common part dominates and the answer does not move when the video changes (89 to 91\% stuck).
Geometry, not information: the bridge input has the same speed consistency in both models (0.139, 0.131), and CF builds a shared speed axis inside the language model ($\kappa$ 0.035\,$\to$\,0.648).
Intervention (\cref{tab:ablate}): removing 8 speed directions (estimated on training clips) takes unseen speed combinations from 98.9 to 65.7 and TOMATO from 50.9 to 46.4; a random subspace of the same energy leaves 97.2 and 49.9, and the top content directions, at 2.6 times the energy, cost speed less and content more.

\begin{table}[t]
\centering\footnotesize\setlength{\tabcolsep}{2.8pt}
\begin{tabular}{lrrrrrr}
\toprule
& & \multicolumn{2}{c}{held-out speed} & & \multicolumn{2}{c}{content} \\
\cmidrule(lr){3-4}\cmidrule(lr){6-7}
Removed subspace & energy & ratio & comb. & TOM & NQA & PT \\
\midrule
\rowcolor{ours}none & 0 & 98.7 & 98.9 & 50.9 & 74.0 & 75.3 \\
speed directions (8) & 16.6 & 91.5 & \textbf{65.7} & \textbf{46.4} & 73.2 & 73.8 \\
random, same energy (24) & 16.2 & 96.7 & 97.2 & 49.9 & 73.2 & 74.3 \\
top content directions (24) & 43.6 & 95.0 & 84.3 & 44.8 & 72.4 & 72.0 \\
\bottomrule
\end{tabular}
\caption{Removing a subspace from every visual token of Ours-2B-Q at inference. Energy: \% removed. Full: \cref{tab:app_ablate}.}
\label{tab:ablate}
\end{table}

\subsection{Explicit pairs versus balanced rows}
\label{sec:pairs}
Two more 2B models draw the same rows per source and answer class from one pool, one using every scene at most once, the other keeping same-scene rows with different answers together (\cref{prop:unpaired}, \cref{app:pairs}). Pairing shows most in consistency: both videos of an unseen speed-ratio pair are correct in 96.7\% against 84.4\%, with gains on state change ($+4.5$) and reversal ($+4.6$).

\section{Discussion and Conclusion}
\label{sec:conclusion}

Captions and questions leave a world-model encoder's dynamics unread. The encoder decides what can be read (direction, counting, fine actions) and the supervision what is read (speed, camera motion, state change), consistent with the measured $\kappa$; with counterfactual supervision, a frozen predictive encoder, one stage and far less data, the model leads every size~class.
Scope and future work. The encoder predicts video rather than reading text, so text recognition and single images are outside our target; hour-long videos, too costly to train on at our scale, are future work.

\iftoggle{vjlmfinal}{%
\section*{Acknowledgements}
This work was supported by Institute of Information \& communications Technology Planning \& Evaluation (IITP, AI Computing Support Project for R\&D) grant funded by the Korea government (MSIT) (No.~RS-2026-25505492, High-Performance Research AI Computing Infrastructure Support at the 2 PFLOPS Scale).
}{}

{
    \small
    \bibliographystyle{ieeenat_fullname}
    \bibliography{main}
}

\clearpage
\onecolumn
\appendix
\section*{Appendix}
\addcontentsline{toc}{section}{Appendix}

\section{Theory: statements and proofs}
\label{app:theory}

\paragraph{Setting.}
$z(v)$ are the frozen encoder tokens of video $v$; a transformation $T$ has an answer map $\pi_T$ with $y(Tv)=\pi_T(y(v))$.
For a binary answer that flips, $y(Tv)=-y(v)$.
The answer log-odds at the scoring position is $s(v)=\ell_P-\ell_Q$, the difference of the two option-letter logits; in the linear regime $s(v)\approx s_0+g\cdot z(v)$ with $g=\partial s/\partial z$ the readout through bridge and language model; we append a constant coordinate to $z$, so that $s_0$ is part of $g\cdot c$.
The pair decomposition $c=\frac12(z(v)+z(Tv))$, $d=\frac12(z(v)-z(Tv))$ is a definition. Here $z(Tv)$ is the input the model actually receives, and $g$ reads tokens at their positions; the encoder analysis of \cref{sec:measure} uses the aligned $z_T(v)=\Pi_T^{-1}z(Tv)$ instead, to separate token content from token position. Since $g\cdot z(Tv)=(\Pi_T^\top g)\cdot z_T(v)$, the two decompositions give the same $m$ and $d_s$ for a readout that weights all token positions alike ($\Pi_T^\top g=g$, \eg through pooling) or when $T$ keeps positions ($\Pi_T=I$); then the subspaces of \cref{thm:a} apply to the pair loss. A position-dependent readout can in addition compare tokens across positions, the relational route of \cref{sec:analysis}, which the propositions do not model; our bridge adds position codes, so the propositions describe the position-invariant part of the readout and the rest is measured (\cref{app:internal}).
The log-odds splits the same way: $m=\frac12(s(v)+s(Tv))=g\cdot c$ and $d_s=\frac12(s(v)-s(Tv))=g\cdot d$.

\paragraph{Proof of \cref{prop:pair}.}
(i) With logistic $\ell(u)=\log(1+e^{-u})$, $\partial_g\ell(y\,s(v))=\ell'(y\,s(v))\,y\,z(v)$ and $\partial_g\ell(-y\,s(Tv))=-\ell'(-y\,s(Tv))\,y\,z(Tv)$.
Writing $a=-\ell'(y\,s(v))$, $b=-\ell'(-y\,s(Tv))$ and substituting $z(v)=c+d$, $z(Tv)=c-d$ gives $\nabla_gL=-y[a(c+d)-b(c-d)]=-y[(a-b)c+(a+b)d]$.
The two margins are $y\,s(v)=y(m+d_s)$ and $-y\,s(Tv)=-y(m-d_s)$, equal exactly when $m=0$, and then $a=b$.
(ii) As a function of $(m,d_s)$, $L=\ell(y d_s+y m)+\ell(y d_s-y m)$. Replacing $m$ by $-m$ swaps the two terms, so $L$ is even in $m$; each term is convex, so $L$ is convex in $m$, and an even convex function is minimized at $m=0$.
If the readout can change $m=g\cdot c$ without changing $d_s=g\cdot d$, which holds when $c$ and $d$ lie in subspaces the readout can address separately (\eg $V_+$ and $V_-$ of \cref{thm:a}), the minimizer over $g$ therefore has $g\cdot c=0$.
\emph{Without separability.} If $c$ and $d$ are collinear, $m$ and $d_s$ move together and $m=0$ is not implied. One-dimensional example: $c=2$, $d=1$, $y=1$, $s=g\,z$, so $L(g)=\ell(3g)+\ell(-g)$; then $L'(0)=-1$ and the minimizer $g^*\approx0.454$ has $m=g^*c\approx0.91\neq0$. This is why we measure the content share at the answer position (\cref{sec:inside}) rather than assume it.
\emph{Answer-balanced sets.} For binary answers and a set of variants $z_j=c+d_j$ of one scene with labels $y_j$ and equal error weights $a_j=a$, the content coefficient of the summed gradient is $-a\,c\sum_jy_j$, which is zero when the labels are balanced. Multi-class answers are treated as pairs of options and are not covered beyond that.

\setcounter{proposition}{3}
\begin{proposition}[unpaired supervision admits shortcuts]
\label{prop:unpaired}
A single clip's gradient is $-a\,y\,(c+d)$; if scene type correlates with the answer ($\mathbb{E}[y\,c]\neq0$), at equal error weights the readout acquires a content component and a counterfactual pair stays where the content puts it.
\end{proposition}

\paragraph{Proof of \cref{prop:unpaired}.}
For a single clip, $\nabla_g\ell(y\,s(v))=-a\,y\,(c+d)$.
Averaged over the training distribution, the content part is $-\mathbb{E}[a\,y\,c]$. At equal error weights (as at the start of training) this is $-a\,\mathbb{E}[y\,c]$, nonzero whenever scene type and answer correlate, so the readout can acquire the component $\mathbb{E}[y\,c]$; later the weights $a$ depend on the scores and the term need not vanish.
On a counterfactual pair the two videos share $c$, so the common term $g\cdot c$ is the same for both and pushes them to the same answer: the pair is ``stuck''.

\begin{proposition}[readability from few pairs, Gaussian model]
\label{prop:kappa}
From a zero readout all pair margins are equal, so the first gradient step on $k$ pairs is the prototype $g_k\propto\sum_i y_i d_i$.
If $d=y\mu+\epsilon$ with $\epsilon\sim\mathcal N(0,\sigma^2 I_D)$ independent of $y$, its accuracy in classifying the change $d$ of a new pair is $\approx\Phi\big(r/\sqrt{r(1+1/k)+D/k}\big)$ with $r=\lVert\mu\rVert^2/\sigma^2=D\kappa/(1-\kappa)$: at fixed $D$ and $k$ it increases with $\kappa$, and it is $\frac12$ at $\kappa=0$.
Applied to a single clip $z=c+d$ (with threshold at the mean content $\bar c$), the decision gains the content term $y\,g_k\cdot(c-\bar c)$, which vanishes when $g_k$ is orthogonal to the content variation and is small when the change direction overlaps little with the leading content directions ($\rho$, \cref{tab:app_enc}).
\end{proposition}

\paragraph{Proof of \cref{prop:kappa}.}
At $g=0$ every margin is zero, so $a=b=-\ell'(0)=\frac12$ for every pair and, by \cref{prop:pair}(i), $\nabla_gL=-\sum_iy_i d_i$: the first gradient step is the prototype $g_k=\frac1k\sum_iy_id_i$ (later steps reweight pairs by their margins, which the proposition does not cover).
Under the Gaussian model, $g_k=\mu+\eta$ with $\eta\sim\mathcal N(0,\frac{\sigma^2}{k}I_D)$, and for the change of a new pair the decision is $y\,g_k\cdot d=(\mu+\eta)\cdot(\mu+\epsilon')$, with mean $\lVert\mu\rVert^2$ and variance $\sigma^2\lVert\mu\rVert^2(1+\frac1k)+D\sigma^4/k$ (the term $\eta\cdot\epsilon'$ dominates at large $D$). Dividing the mean by the standard deviation gives the stated $\Phi\big(r/\sqrt{r(1+1/k)+D/k}\big)$, which tends to the Bayes accuracy $\Phi(\sqrt r)$ as $k\to\infty$; this is a normal approximation of the decision, which in simulation ($k=8$, $r\le16$) is within 0.4 points for $D\ge64$, the regime of the encoders, and within 2.5 points down to $D=1$. With $\kappa=\lVert\mu\rVert^2/(\lVert\mu\rVert^2+D\sigma^2)$, $r=D\kappa/(1-\kappa)$.
At $\kappa=0$, $\mu=0$ and the decision is symmetric about zero, so the accuracy is $\frac12$. Symmetry matters: with a skewed residual, $\kappa=0$ does not imply chance. The statement is about this linear estimator; a nonlinear readout of relations between tokens is not covered.

\paragraph{Supervision and the answer layers (stated in \cref{sec:method}).}
At equal margins the pair gradient of the readout lies in $\mathrm{span}\{d\}$ (\cref{prop:pair}(i)). This is a statement about the readout, not about every update of the language model; whether training raises the cross-video alignment $\kappa_L$ inside the model is measured, not derived (\cref{app:internal}).

\paragraph{Proof of \cref{thm:a}.}
If the loss satisfies $\mathcal{L}(E\circ T)=\mathcal{L}(\Pi_T\circ E)$ and $E$ is a minimizer, then $\Pi_T^{-1}\circ E\circ T$ is also a minimizer.
By the identifiability assumption, $\Pi_T^{-1}E(Tv)=A_TE(v)$ for an invertible $A_T$ that acts on token channels, the same at every position, so it commutes with every token permutation.
Applying the identity twice gives $A_{T_1T_2}=A_{T_1}A_{T_2}$, a representation.
For $T^2=1$, $A_T^2=I$ and $A_T$ is diagonalizable with eigenvalues $\pm1$, so $\mathbb{R}^D=V_+\oplus V_-$.
Then $c=P_+z\in V_+$ and $d=P_-z\in V_-$ for every $v$.
\emph{Readout.} Let $y=\mathrm{sign}(w\cdot z)$ with $y(Tv)=-y(v)$ for all $v$, and let $w$ weight all token positions alike ($\Pi_T^\top w=w$), so that $w\cdot z(Tv)=w\cdot A_Tz(v)$; a position-dependent readout can instead compare tokens across positions (the relational route). With $A=A_T$ this reads $\mathrm{sign}(w\cdot Az)=-\mathrm{sign}(w\cdot z)$.
Then $w_-\cdot z=\frac12(w\cdot z-w\cdot Az)$ is a sum of two terms that both have the sign of $y$, so $w_-$ gives the same answer on every video; and since $A^\top w_-=-w_-$, $w_-\cdot c=w_-\cdot\frac12(I+A)z=\frac12\big[(I+A^\top)w_-\big]\cdot z=0$.
A sign flip alone does not give $w\cdot Az=-w\cdot z$ pointwise, so $w$ itself may carry content when the data avoid the boundary. If the relation holds on an open set that meets $\{w\cdot z=0\}$, the two linear forms $w\cdot z$ and $(A^\top w)\cdot z$ vanish on the same hyperplane there with opposite signs, so $A^\top w=-\lambda w$ with $\lambda>0$; $(A^\top)^2=I$ gives $\lambda=1$, and $w=w_-$.
The readout lies in the $-1$ eigenspace of $A^\top$, which annihilates $V_+$ (it coincides with $V_-$ when $A$ is orthogonal).
\emph{Speed.} If playback rates act as a one-parameter group $\tau_a$, the same argument gives $A_{\tau_a}=\exp(aK)$, and a small rate change moves the features by $aKz$. This direction depends on $z$, so the proposition does not give speed a direction shared by all videos (it would for a rank-one $K$, which nothing forces); how consistent the speed change is in a real encoder is an empirical question, answered by $\kappa$ in \cref{tab:app_enc}.
\emph{Order.} For tokens that see only a temporally local window, a permutation of segments longer than that window acts on tokens by $\Pi$ only ($A=I$); for encoders with temporal context across segments it also changes token content near the boundaries.
\emph{Composition.} If mirror $M$ and reversal $R$ both act as $-1$ on the motion subspace, then $A_{MR}=A_MA_R=+1$ there, so the composed video reads like the original. We test this in \cref{app:internal}.
\emph{Scope of the assumptions.} For V-JEPA the symmetry holds approximately: video and masks are left--right symmetric, while masks are tubes through time and the frame stride is random. \Cref{sec:measure} therefore tests the consequences, $\kappa$ and $\rho$, on the frozen features directly.

\paragraph{Proof of \cref{thm:b}.}
(i) If $E(v)=[f(x_1),\dots,f(x_N)]$, time reversal maps $E(v)$ to $[f(x_N),\dots,f(x_1)]=\Pi E(v)$, so $\Pi^{-1}E(Tv)=E(v)$ and $d\equiv0$. For our SigLIP tokens, which pack two consecutive frames, $\Pi$ also swaps the two frames inside each token. Without that swap, as the model receives them, $d=\frac12[f(x_{2t-1})-f(x_{2t})]$ per frame block, a scene-dependent frame difference ($r_d=0.055$, $\kappa=0.0012$ measured). This is a statement about token content after the positions are aligned: the order of the tokens still differs between $v$ and $Tv$, and a model that compares tokens across positions can recover the arrow of time (the frozen SigLIP probe reads it at 65.3\%, \cref{app:enc}).
A change of rate selects other frames, $d=\frac12[f(x_{t})-f(x_{t'})]$, a frame difference that depends on scene content; nothing in the objective aligns it across videos, so a shared direction is not implied (measured: $\kappa\approx0.002$, \cref{tab:app_enc}).
(ii) The contrastive loss depends on $x$ only through $f(x)\cdot g(\text{caption})$.
If captions are invariant to mirroring in distribution, the expected loss is the same for $f$ and $f\circ\text{mirror}$. For unnormalized embeddings the contrastive loss is convex in $f$ (a log-sum-exp of linear terms), so $\frac12(f+f\circ\text{mirror})$ does no worse on the caption-aligned subspace. For normalized embeddings the average leaves the sphere and the argument does not apply as stated; there we rely on the measured mirror consistency ($\kappa=0.0014$ for SigLIP, \cref{tab:app_enc}).
A minimizer can therefore be mirror-invariant there, and the remaining mirror component is unconstrained residual.

\FloatBarrier
\section{Counterfactual data}
\label{app:data}
\paragraph{Relation to the closest work.} \Cref{tab:app_related} sets the counterfactual rows against the closest training signals and analyses.
\begin{table*}[!htbp]
\centering\footnotesize\setlength{\tabcolsep}{3pt}
\begin{tabular}{@{}p{2.9cm}p{3.8cm}p{3.4cm}p{2.8cm}p{3.5cm}@{}}
\toprule
Work & Training signal & Where the answer comes from & Encoder & What is analysed \\
\midrule
Pretext tasks~\cite{misra2016shuffle,fernando2017odd,wei2018arrow,xu2019clip,benaim2020speednet,yao2020prp,wang2020pace} & order, arrow-of-time or speed prediction & the transformation & trained by the task & learned features \\
Tracing the Arrow of Time~\cite{han2026tracing} & arrow-of-time questions, time-preserving connector & reversal & video- vs.\ frame-centric & information flow, encoder to answer \\
Which Way Did It Move?~\cite{lee2026whichway} & projector objective on frame-feature deltas (DeltaDirect), direction instructions & 2-D motion direction & -- & direction readable at each stage, binding gap, direction concept vectors \\
TIME~\cite{wang2026time} & answer-balanced temporal instructions, five dimensions & constructed instructions; reversal for direction & -- & benchmark gains \\
SynRL~\cite{jiang2026synrl} & CoT fine-tuning and RL on rendered shapes & the rendering program & -- & synthetic-to-real transfer \\
\rowcolor{ours}Ours & 14 constructions in 6 families on public videos, added to captions and QA & the transformation of the same clip, so content cancels & frozen predictive vs.\ frame-wise & change structure ($\kappa$, $\rho$), stage-wise probes, subspace removal, encoder $\times$ supervision \\
\bottomrule
\end{tabular}
\caption{The counterfactual rows and the closest training signals and analyses.}
\label{tab:app_related}
\end{table*}

\paragraph{Adding a video collection.} A new source is a short script: it lists the collection's clips (path and, for constructions on actions and events, the source's own action label or time span) and names the constructions to apply. The generator then renders the variants, writes question, options and answer, and applies the answer, option-length and text-prior balancing of \cref{sec:method}; no manual annotation and no language model are involved, so any public video collection becomes counterfactual training data.

\begin{figure*}[!htbp]
\centering
\includegraphics[width=\textwidth]{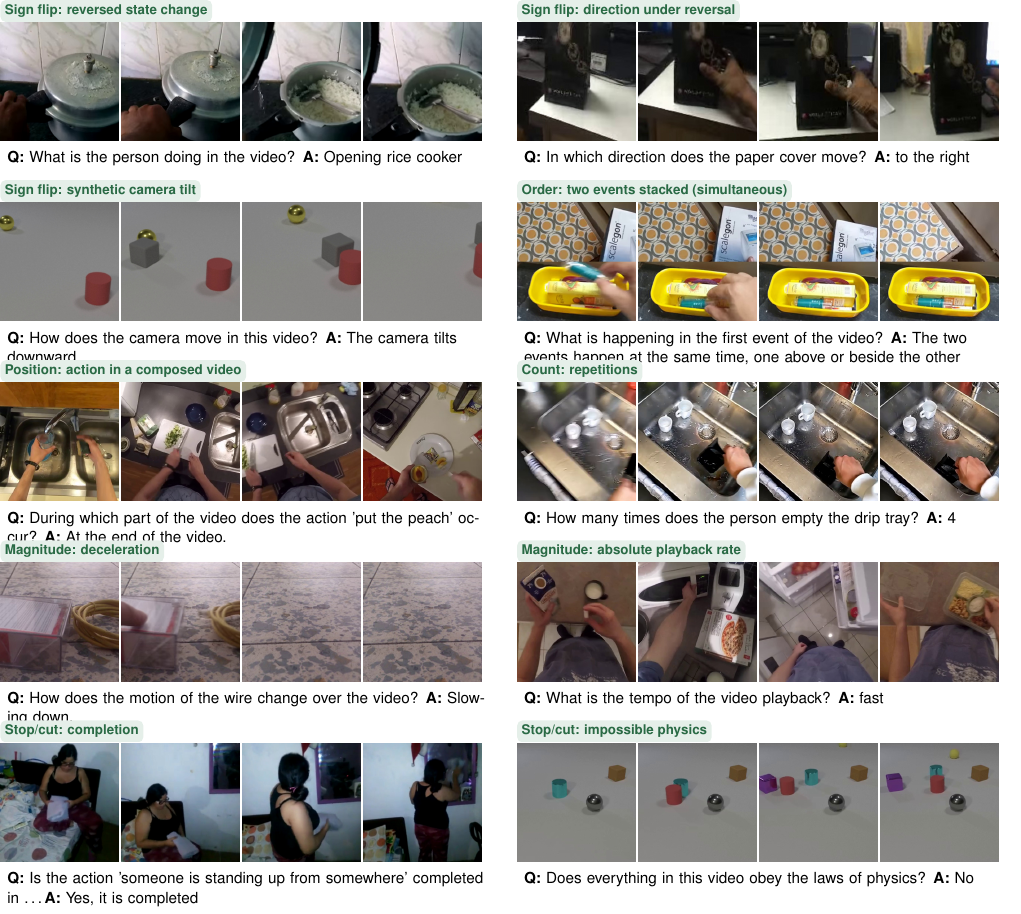}
\caption{One training row per construction (4 of 32 frames, after the transformation), with its question and answer.}
\label{fig:app_examples}
\end{figure*}

\Cref{tab:app_data} lists all 33 sources with their construction, size and a real question.
\Cref{fig:app_examples} shows one rendered training row per construction.
The sources draw on SSv2, SA-V, Charades-Ego, EPIC-Kitchens, HoloAssist, Kinetics-700, CLEVRER and a physics simulator.
Each row is rendered at training time from its specification (clip, window, transformation), so the same file yields the original and its counterfactual without storing extra video.
\Cref{fig:example} in the main text is a sign-flip row (state change played backwards).

\paragraph{How each construction is rendered.}
All constructions are rendered on the fly by the training loader from a source clip (and, where noted, a second clip), at 32 frames, with frame times replaced by even spacing so that the answer cannot be read off time codes.
\begin{enumerate}[leftmargin=*,itemsep=1pt,topsep=2pt,label=\textbf{C\arabic*.}]
\item \textbf{State change played backwards} (sign flip). Actions with an inverse (open/close, put/take, turn on/off) from SSv2, Charades-Ego and EPIC; the clip is played forward or reversed, and the answer becomes the inverse action when reversed. A variant asks for the end state.
\item \textbf{Object direction under mirroring or reversal} (sign flip). SSv2 clips with a direction in the template (left/right, up/down, towards/away). The clip is mirrored (left$\leftrightarrow$right) or reversed (all directions flip), and the answer is remapped accordingly.
\item \textbf{Camera move} (sign flip). A square window of 0.7 of the short side moves over static footage (CLEVRER, Charades-Ego): pan left/right, tilt up/down, zoom in (window shrinks to 60\%) or out, or stays static. The amount travelled is drawn per row, and SSv2 camera-direction clips are additionally mirrored or reversed.
\item \textbf{Two clips joined, each optionally reversed} (sign flip). The first and second halves come from two SSv2 clips; the answer is ``direction A then direction B''.
\item \textbf{Event order} (order). Two clips are joined in sequence or stacked above and below in the same time span (simultaneous). The answer is first, second, or at the same time, and natural clips with annotated event order are added. For simultaneous rows the layout itself shows both events at once; the order answers (first or second) are the ones only the change decides.
\item \textbf{Caption--segment matching} (order). Two HoloAssist segments are separated by black frames, and a caption names two actions in order; the matching segment is the answer.
\item \textbf{Temporal position} (position). Three clips are joined; the queried action sits in one, and the answer is beginning, middle or end. Variants place two Kinetics clips around a black gap and ask what happens in a part or which part shows an action.
\item \textbf{Relative speed} (magnitude). One clip is shown twice, above and below, with the frames of each copy spread evenly over a span $[a,b]$ of the clip; the copy with the longer span moves faster. The spans are nested, the shorter one centred inside the longer one, so both copies show the same motion and the longer span plays more of it in the same number of frames; the audit below checks that apparent motion follows the label on every faster and slower row. Swapping the spans flips the answer, and equal spans give ``the same''. Span ratios are 2:1, 3:1 or equal; the fine-motion source compares spans with ratios between 1.0 and 1.4.
\item \textbf{Speed change} (magnitude). The time span is split 1:3, 3:1 or 1:1 between the two halves of the frames, so the second half moves three times as fast, a third as fast, or as fast as the first.
\item \textbf{Playback rate} (magnitude). The frames are spread over a short, normal or long window: slow motion, normal or time-lapse.
\item \textbf{Repetitions} (count). One action segment is repeated $k$ times ($k\leq7$), and the answer is $k$.
\item \textbf{Distinct activities} (count). One to four segments of different Kinetics activities are joined, and the answer is their number.
\item \textbf{Completion} (stop/cut). The clip is cut before or after the end of an annotated action, and the answer is complete or incomplete.
\item \textbf{Physical plausibility} (stop/cut). Simulated scenes are rendered with and without an impossible edit (teleport, vanishing object, \dots); both versions are in the data, so every row is paired.
\end{enumerate}
Across all constructions the generator balances answer letters, the rank of the correct option by length, and each option's prior under the text-only language model.

\begin{table*}[!htbp]
\centering\scriptsize\setlength{\tabcolsep}{3pt}
\begin{tabular}{>{\raggedright\arraybackslash}p{3.2cm}>{\raggedright\arraybackslash}p{5.0cm}rrr>{\raggedright\arraybackslash}p{5.3cm}}
\toprule
Source & Construction & Rows & \% mix & paired \% & Example question (real row) \\
\midrule
\multicolumn{6}{l}{\textit{Sign flip} (12 sources, 117,066 rows, 1.47\% of the mixture)} \\
tq4\_odir\_ssv2 & object direction; clip mirrored or reversed with the answer remapped & 16,116 & 0.15 & 0 & In which direction does the roll move? \\
tq4\_cdir\_ssv2 & camera direction under mirroring / reversal & 5,895 & 0.05 & 0 & What is the camera motion in the video? \\
tq4\_dir2\_ssv2 & two concatenated clips, each optionally reversed: direction A then B & 7,204 & 0.07 & 0 & How would you describe the bowl's directional movement? \\
tq4\_state\_ssv2 & state-change action reversed: open/close, put/take & 24,659 & 0.23 & 0 & Which action is performed in the video? \\
chego\_rev & egocentric actions with reversible verbs, clip reversed & 16,780 & 0.26 & 7 & Which action does the person perform? \\
epic\_rev & kitchen actions with reversible verbs (on/off, open/close) & 3,406 & 0.05 & 0 & What am I doing in this video? \\
chego\_state & end state of a reversible action (open/closed) & 4,383 & 0.07 & 4 & What state is the closet / cabinet in at the end? \\
camera\_clevrer & synthetic camera move (pan, tilt, zoom, static) over a static scene & 5,000 & 0.08 & 0 & How does the camera move in this video? \\
camera\_chexo & as above, egocentric source & 3,927 & 0.06 & 0 & Which camera movement is shown? \\
camera\_chexo\_q2 & camera move, second question template & 3,668 & 0.06 & 0 & How does the camera move as the scene unfolds? \\
camera\_clevrer\_q2 & camera move, second question template & 4,662 & 0.07 & 0 & Is there any camera movement in the video? \\
v17\_finemotion\_dir\_\_ssv2 & fine motion direction of object or camera & 21,366 & 0.33 & 0 & Is the camera panning to the left? \\
\midrule
\multicolumn{6}{l}{\textit{Order} (4 sources, 85,579 rows, 0.92\% of the mixture)} \\
tq4\_order\_ssv2 & two events concatenated or stacked: which first / simultaneous & 29,648 & 0.28 & 0 & What is happening first in the video? \\
tq4\_order\_llava & as above, open-domain clips & 20,000 & 0.19 & 0 & What is happening in the first event of the video? \\
tq4\_order\_nat & natural event order within one clip & 15,661 & 0.15 & 0 & What order do the events occur in? \\
order\_capmatch & two clips joined; a caption names two actions in order: which clip it describes & 20,270 & 0.31 & 80 & The video has two segments separated by black frames. Which \dots \\
\midrule
\multicolumn{6}{l}{\textit{Position} (4 sources, 48,386 rows, 0.74\% of the mixture)} \\
loc\_comp & three clips joined: where does the action occur & 23,700 & 0.36 & 90 & During which part of the video does the action 'wash the \dots \\
v15\_half\_k700\_act & two clips around a black gap: what happens in the first/second part & 5,972 & 0.09 & 5 & What happens in the first part of the video (after / before \dots \\
v15\_whichseg\_k700\_act & which of two segments shows an action & 5,962 & 0.09 & 5 & The video has two segments separated by black frames. Which \dots \\
loc\_comp\_q2 & localization, second question template & 12,752 & 0.20 & 86 & When does the action 'wash the knife' occur in the video? \\
\midrule
\multicolumn{6}{l}{\textit{Magnitude} (8 sources, 141,084 rows, 1.58\% of the mixture)} \\
tq4\_rspeed\_ssv2 & one clip twice, above/below, over different spans: relative speed & 30,000 & 0.28 & 0 & Compared to the video below, how fast is the tea bag above \dots \\
tq4\_rspeed\_sav & as above, SA-V source & 18,000 & 0.17 & 0 & What is the speed of the video above compared to the video \dots \\
tq4\_dspeed\_ssv2 & span split 1:3 / 3:1 / 1:1: accelerates, decelerates, constant & 30,000 & 0.28 & 0 & What is the empty plastic container's speed pattern? \\
tq4\_dspeed\_sav & as above, SA-V source & 18,000 & 0.17 & 0 & How does the speed of the motion change? \\
v16\_rspeed\_w\_\_tq4\_rspeed\_sav & relative speed (2:1 / 3:1 / equal), varied question wording & 7,926 & 0.12 & 0 & What is the speed of the video above? \\
v16\_rspeed\_w\_\_tq4\_rspeed\_ssv2 & relative speed (2:1 / 3:1 / equal), varied question wording & 7,968 & 0.12 & 0 & What is the speed of the bottle above? \\
v16\_tempo\_\_epic & absolute playback rate: slow motion / normal / time-lapse & 11,790 & 0.18 & 0 & What is the relative speed of the video compared to real-time? \\
v17\_finemotion\_speed\_\_ssv2 & fine relative speed of the same object & 17,400 & 0.27 & 0 & How does the speed of the men's inner vest above compare to \dots \\
\midrule
\multicolumn{6}{l}{\textit{Count} (3 sources, 35,614 rows, 0.55\% of the mixture)} \\
rep\_loop & one action segment repeated k times: how many times & 12,000 & 0.18 & 86 & How many times does the person load the inside water \dots \\
v15\_ndistinct\_k700\_act & segments of 1-4 distinct activities: how many & 11,776 & 0.18 & 100 & How many distinct activities appear in this video? \\
v16\_rep6\_\_rep\_loop & repetitions up to 7 & 11,838 & 0.18 & 86 & How many times does the person insert the hex socket head? \\
\midrule
\multicolumn{6}{l}{\textit{Stop/cut} (2 sources, 81,464 rows, 0.85\% of the mixture)} \\
plausibility & physically impossible edit (teleport, vanish, ...) vs. its unedited render & 65,800 & 0.61 & 100 & Is the motion of the objects in this video physically \dots \\
chego\_done & clip truncated before the action completes & 15,664 & 0.24 & 5 & By the end of the clip, has the person finished opening a \dots \\
\bottomrule
\end{tabular}
\caption{All 33 counterfactual sources of the training mixture (509,193 rows, 6.11\% of sampled rows). Paired \%: share of a source's rows whose counterpart from the same clip with a different answer is also in that source (29.4\% overall); the other rows are single members of answer-balanced constructions (\cref{sec:method}).}
\label{tab:app_data}
\end{table*}

\paragraph{Label audit for speed.} The speed constructions manipulate playback, and the questions ask about apparent motion. On 300 random training rows of each SSv2 speed source, rendered by the training loader, we use the mean absolute difference between consecutive frames as a proxy for apparent motion (top versus bottom panel; second versus first half). For relative speed it points the way of the label in 100\% of ``faster'' and ``slower'' rows (median motion ratio 1.83 and 0.52), and 92.5\% of ``same'' rows fall within a ratio of 0.8--1.25; for speed change it agrees on 97.7\% of accelerating and 95.0\% of decelerating rows (median ratios 2.53 and 0.39). Speed-change labels assume the source moves at a roughly steady rate over the clip; the audit confirms the label on 95--98\% of these rows. Relative-speed labels are the playback-rate ratio the rendering sets for two playbacks of the same clip, and apparent motion follows them on every audited faster and slower row; absolute-rate rows (slow motion / normal / time-lapse) take the source as recorded at natural speed. The readout carries over to speed items that external benchmarks build with their own pipelines: there CF adds 32.8 points at 8k and 38.1 at the end (\cref{tab:app_skills}).

\FloatBarrier
\section{Training and evaluation details}
\label{app:train}
\begin{table*}[!htbp]
\centering\footnotesize\setlength{\tabcolsep}{5pt}
\begin{tabular}{llllrrrrr}
\toprule
Model & Language model & Encoder & Batch/GPU$\times$GPUs & Global & Steps & Samples & LR (LLM) & EFLOPs \\
\midrule
Ours-2B-Q & Qwen3-1.7B & ViT-L & 12$\times$14 & 168 & 27,382 & 4.60M & 1.05e-5 & 121 \\
Ours-1B-L & Llama-3.2-1B-Instr. & ViT-L & 6$\times$28 & 168 & 27,382 & 4.60M & 1.05e-5 & 104 \\
Ours-4B-Q & Qwen3-4B-Instr.-2507 & ViT-g & 5$\times$30 & 150 & 30,187 & 4.53M & 1.2e-5 & 314 \\
Ours-3B-L & Llama-3.2-3B-Instr. & ViT-g & 10$\times$30 & 300 & 15,094 & 4.53M & 1.7e-5 & 277 \\
Ours-8B-Q & Qwen3-8B & ViT-g & 5$\times$30 & 150 & 30,187 & 4.53M & 8e-6 & 437 \\
Ours-8B-L & Llama-3.1-8B-Instr. & ViT-g & 5$\times$30 & 150 & 30,187 & 4.53M & 8e-6 & 426 \\
\bottomrule
\end{tabular}
\caption{Training settings. All: frozen V-JEPA~2.1 encoder at 32 frames, spectral bridge to 1,024 tokens (fixed PCA rotation and rescaling, then a trainable MLP), bridge learning rate $2\times10^{-4}$ (3B: $2.83\times10^{-4}$), weight decay 0.01 (LLM) / 0.05 (bridge), linear warm-up over 0.2--0.3\% of steps with the LLM frozen for the first 0.6--0.8\%, cosine decay, maximum sequence 2,560 tokens, 4.53M rows drawn at fixed per-source weights from the 175-source mixture (4.60M for the ViT-L models). Compute: \cref{tab:app_train_stats}; EFLOPs ($10^{18}$): trained parameters $\times$ trained tokens $\times$ 6 plus attention and the frozen encoder's forward pass, measured per clip (activation recomputation excluded).}
\label{tab:app_train}
\end{table*}

\paragraph{Released models.} The six models are released as Llama-V-JEPA-LM-Qwen3-1.7B, -Qwen3-4B, -Qwen3-8B, -Llama3.2-1B, -Llama3.2-3B and -Llama3.1-8B (Ours-2B-Q, -4B-Q, -8B-Q, -1B-L, -3B-L, -8B-L), with weights, inference code and model cards (\url{https://huggingface.co/collections/kkt94/llama-v-jepa-lm-6af41fb2a548b444c6255d63}); code and the technical report are at \url{https://github.com/kkt94/Llama-V-JEPA-LM}.

\begin{figure*}[!htbp]
\centering
\includegraphics[width=\textwidth]{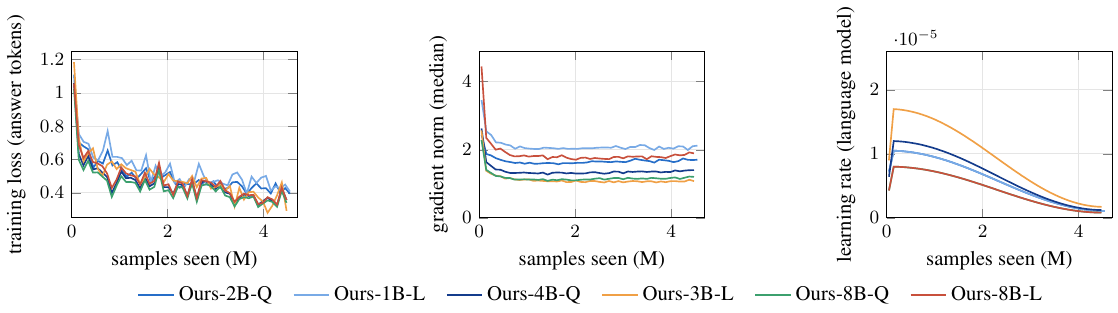}
\caption{Training loss, gradient norm (bin median) and language-model learning rate of the released models against samples seen.}
\label{fig:app_train_curves}
\end{figure*}

\paragraph{Training curves and metrics.} \Cref{fig:app_train_curves} plots the training loss, the gradient norm and the language model's learning rate of the released models against samples seen, and \cref{tab:app_train_stats} lists throughput, hours, memory and loss. Each model is trained once, in one continuous stage. The loss falls steadily without divergence and ends lower for larger language models, from 0.43 (1B) to 0.38 (8B); the Qwen3 and Llama models of the same size end within 0.02 of each other. The language model's learning rate is zero during the bridge-only warm-up, then decays by cosine.
\begin{table*}[!htbp]
\centering\footnotesize\setlength{\tabcolsep}{4.5pt}
\begin{tabular}{lrrrrrrrrrr}
\toprule
Model & H100s & Global batch & Steps & Samples & Samples/s & Hours & H100-hours & Peak mem.\ (GB) & Loss, first 1\% & Loss, last 5\% \\
\midrule
\rowcolor{ours}Ours-2B-Q & 14 & 168 & 27,382 & 4.60M & 78 & 16.4 & 229 & 46 & 1.53 & 0.41 \\
\rowcolor{ours}Ours-1B-L & 28 & 168 & 27,382 & 4.60M & 118 & 10.8 & 303 & 18 & 1.31 & 0.43 \\
\rowcolor{ours}Ours-4B-Q & 30 & 150 & 30,187 & 4.53M & 76 & 16.6 & 499 & 44 & 1.41 & 0.39 \\
\rowcolor{ours}Ours-3B-L & 30 & 300 & 15,094 & 4.53M & 88 & 14.3 & 429 & 51 & 1.51 & 0.39 \\
\rowcolor{ours}Ours-8B-Q & 30 & 150 & 30,187 & 4.53M & 58 & 21.6 & 648 & 33 & 1.36 & 0.38 \\
\rowcolor{ours}Ours-8B-L & 30 & 150 & 30,187 & 4.53M & 63 & 19.9 & 596 & 32 & 1.37 & 0.40 \\
\bottomrule
\end{tabular}
\caption{Training metrics of the released models (one stage, one run each). Samples/s: median throughput over the run; hours and H100-hours: samples divided by that throughput, so pauses between resumed segments are not counted. Peak memory: per GPU (activation checkpointing). Loss: mean next-token cross-entropy on answer tokens of the logged batches.}
\label{tab:app_train_stats}
\end{table*}

\begin{table}[!htbp]
\centering\footnotesize\setlength{\tabcolsep}{2.6pt}
\begin{tabular}{lrrrrrrrr}
\toprule
Samples (M) & 2.7 & 3.0 & 3.3 & 3.6 & 3.9 & 4.2 & 4.5 & 4.53 \\
\midrule
\rowcolor{ours}Ours-4B-Q & 69.97 & 70.24 & 69.98 & 71.43 & 71.41 & 71.73 & 72.18 & 72.14 \\
\rowcolor{ours}Ours-3B-L & 66.40 & 67.14 & 67.39 & 68.33 & 68.73 & 68.54 & 68.92 & 68.97 \\
\rowcolor{ours}Ours-8B-Q & 68.66 & 68.47 & 69.28 & 69.89 & 70.12 & 69.92 & 70.23 & 70.11 \\
\rowcolor{ours}Ours-8B-L & 68.40 & 68.82 & 68.79 & 69.78 & 70.41 & 70.59 & 70.61 & 70.51 \\
\bottomrule
\end{tabular}
\caption{Twelve-score average at matched training samples for the ViT-g models (3B: global batch 300; 4B, 8B: 150). Each model is first in its class at every point. The ViT-L models (1B, 2B; global batch 168) are checkpointed at other sample counts; the 2B trajectory is in \cref{fig:traj}.}
\label{tab:sizes}
\end{table}

\paragraph{Spectral bridge.}
The frozen encoder's states are layer-normalized per level, pooled from the $24{\times}24$ patch grid to $8{\times}8$ cells and kept per tubelet, so 32 frames give 16 tubelets $\times$ 64 cells $=$ 1,024 tokens.
Each cell state is centered, rotated to its principal axes and scaled by one global factor to unit mean variance; the statistics are computed once on 2.3M cell states from 4,500 training-mix clips and then fixed. ViT-L (1B, 2B) keeps all 1,024 directions; ViT-g (3B--8B; four levels, 5,632 dimensions) keeps the leading 2,048.
A trainable MLP (LayerNorm, linear to 2,048, GELU, linear to the language-model width) maps each cell to a token, and learned codes for tubelet, row, column, time stamp and cell geometry are added.
The SigLIP arm uses the same pipeline: each frame is encoded on its own, and the two frames of a V-JEPA tubelet form the two levels of one token (2,304 dimensions, leading 2,048 kept), so frames, token layout and token count are identical.

\paragraph{Mixture.}
The 175 sources comprise open video caption and QA collections---LLaVA-Video~\cite{zhang2025llavavideo}, ShareGPTVideo (LLaVA-Hound)~\cite{zhang2024llavahound}, PerceptionLM data~\cite{cho2025plm}, the VideoChat-Flash and TimeIT mixtures~\cite{li2024videochatflash,ren2024timechat}, ActivityNet-QA and ActivityNet Captions~\cite{yu2019activitynetqa,krishna2017densecap}, STAR~\cite{wu2021star}, VidLN~\cite{voigtlaender2023vidln} and the PerceptionTest training split~\cite{patraucean2023perception}---; questions built from the original labels of SSv2~\cite{goyal2017ssv2}, Kinetics-700~\cite{carreira2019kinetics700}, Charades-Ego~\cite{sigurdsson2018charadesego}, EPIC-KITCHENS-100~\cite{damen2022epic100}, HoloAssist~\cite{wang2023holoassist}, EgoDex~\cite{hoque2026egodex}, SA-V~\cite{ravi2025sam2}, CLEVRER~\cite{yi2020clevrer}, RoboVQA~\cite{sermanet2024robovqa}, Language-Table~\cite{lynch2023interactive}, Oops!~\cite{epstein2020oops}, Countix~\cite{dwibedi2020repnet} and HT-Step~\cite{afouras2023htstep}; text-only replay on UltraChat prompts~\cite{ding2023ultrachat}; and the 33 counterfactual sources. Every source is used through its training split. Video encoders and language models are used as released~\cite{murlabadia2026vjepa21,zhai2023siglip,qwen3,llama3}.
Sampling weights fix each source's share of a step: the 4.53M drawn rows come from 7.2M unique rows (a median source row is drawn 0.7 times), and the 509k counterfactual rows give about 277k draws.

\paragraph{Evaluation.}
Every model, ours and open, is evaluated by the same harness: 32 uniformly sampled frames, the benchmark's question and options with the instruction ``Answer with the option's letter from the given choices directly'', and scoring on the option letter.
Training samples in \cref{fig:efficiency} are as each paper reports them: InternVL3.5 116M pre-training + 56M SFT~\cite{internvl35}; PLM 1M + 72M + 19M~\cite{cho2025plm}; VideoLLaMA3 the sum of its four stage tables~\cite{videollama3}; Molmo2 its SFT mixture, since its image pre-training count is not reported, so its 9.3M is a lower bound~\cite{molmo2}; Qwen3-VL reports 2.2T pre-training tokens~\cite{qwen3vl} and is not plotted (58.4), and Meta's V-JEPA~2 LLM uses 88.5M samples (\cref{tab:meta}).
Open models use their released instruction-tuned weights from the Hugging Face Hub, their own processors and their own chat templates at 32 frames, with thinking off where a template offers it (Gemma~4); our models use their chat template without a thinking block. Scoring takes the next-token logits at the first assistant position and picks the option letter with the highest logit among the item's options, with no generation and no answer parsing, so every model gives a valid answer for every item. As a check, greedy generation (16 tokens) with answer parsing gives the same prediction as the letter logits on 100\% of TOMATO and MVBench items for Qwen3-VL-4B, Qwen3-VL-8B, InternVL3.5-4B and PLM-3B and on 98.7--99.5\% for Ours-4B-Q and Ours-8B-Q, with no unparsed answer; the four open models' scores are unchanged and ours move by at most 0.3 (TOMATO 53.2 and 56.6, MVBench 72.0 and 72.8). \Cref{tab:app_reported} sets the numbers the authors report, under their own frame settings and splits, beside ours: of 51 reported scores, 40 lie within 3 points of our measurement (median difference 1.3). Resampling the items of every benchmark (whole pairs for pair metrics; 1,000 draws, the same draws for every model) gives 95\% intervals of the 12-score average of $\pm$0.68--0.75 points for every model in \cref{tab:main}, so the margins over the best open model of each class (5.3 to 13.6 points) lie far outside them.
\begin{table*}[!htbp]
\centering\footnotesize\setlength{\tabcolsep}{4pt}
\begin{tabular}{lrrrrrrr}
\toprule
Model & MVBench & PerceptionTest & TVBench & TempCompass & TOMATO & MotionBench & NExT-QA \\
\midrule
\multicolumn{8}{@{}l}{\emph{2B class}} \\
Qwen3-VL-2B & 61.7 (61.0) & -- (65.1) & -- (49.8) & -- (67.4) & -- (30.3) & -- (57.5) & -- (77.1) \\
InternVL3.5-2B & 65.9 (66.8) & -- (64.2) & -- (50.9) & -- (65.7) & -- (31.0) & -- (56.2) & -- (76.1) \\
PLM-1B & 70.1 (68.0) & 72.7$^t$ (69.8) & 50.4 (48.2) & 64.6 (60.5) & 25.5 (24.5) & 52.2 (51.7) & 80.3 (79.8) \\
VideoLLaMA3-2B & 65.5 (57.9) & 68.0$^t$ (62.9) & -- (44.0) & 63.4 (64.1) & -- (24.0) & -- (56.3) & 81.1 (75.4) \\
Apollo-1.5B & -- (61.1) & 61.0 (63.9) & -- (47.9) & 60.8 (60.3) & -- (27.8) & -- (51.1) & -- (78.8) \\
\rowcolor{ours}Ours-2B-Q & 69.5 & 75.3 & 70.7 & 73.5 & 50.9 & 54.3 & 74.0 \\
\multicolumn{8}{@{}l}{\emph{4B class}} \\
Qwen3-VL-4B & 68.9 (68.0) & -- (72.0) & -- (59.2) & -- (70.9) & -- (36.3) & -- (61.5) & -- (80.9) \\
InternVL3.5-4B & 71.2 (70.9) & -- (69.1) & -- (60.0) & -- (71.1) & -- (32.1) & -- (59.0) & -- (79.8) \\
PLM-3B & 74.7 (73.0) & 79.3$^t$ (76.0) & 58.9 (56.6) & 69.3 (67.0) & 30.9 (36.3) & 60.4 (58.9) & 83.4 (84.0) \\
Molmo2-4B & 75.1 (74.2) & 81.3$^t$ (81.4) & -- (65.5) & 72.8 (70.8) & 39.8 (36.5) & 61.6 (62.3) & 85.5 (85.1) \\
Apollo-3B & -- (62.0) & 65.0 (67.1) & -- (50.9) & 62.5 (63.0) & -- (29.0) & -- (55.0) & -- (80.9) \\
\rowcolor{ours}Ours-4B-Q & 72.1 & 78.2 & 74.8 & 79.9 & 53.5 & 59.1 & 79.3 \\
\multicolumn{8}{@{}l}{\emph{8B class}} \\
Qwen3-VL-8B & 68.7 (68.8) & -- (73.0) & -- (58.2) & -- (73.6) & -- (36.5) & -- (62.9) & -- (82.1) \\
InternVL3.5-8B & 72.1 (71.0) & -- (72.7) & -- (62.7) & -- (71.0) & -- (33.6) & -- (58.1) & -- (81.5) \\
PLM-8B & 77.1 (76.0) & 82.7$^t$ (80.6) & 63.5 (62.7) & 72.7 (70.9) & 33.2 (42.6) & 61.4 (62.7) & 84.1 (85.7) \\
Molmo2-8B & 75.9 (75.6) & 82.1$^t$ (81.9) & -- (64.1) & 73.4 (72.1) & 39.6 (35.4) & 62.2 (61.4) & 86.2 (85.4) \\
VideoLLaMA3-7B & 69.7 (65.6) & 72.8$^t$ (69.7) & -- (49.2) & 68.1 (69.3) & -- (29.4) & -- (57.8) & 84.5 (81.5) \\
\rowcolor{ours}Ours-8B-Q & 72.8 & 77.8 & 75.3 & 79.4 & 56.5 & 59.1 & 81.2 \\
\bottomrule
\end{tabular}
\caption{Scores reported by the models' authors, with ours under the common protocol in parentheses (32 frames, option-letter scoring; \cref{tab:main}). Reported settings differ from ours: PerceptionTest test split ($^t$; ours: validation), Qwen3-VL at 2 fps up to 2,048 frames, InternVL3.5 the best of 16--64 frames, Molmo2 at 2 fps up to 128 frames, VideoLLaMA3 at 1 fps up to 180 frames, PLM 32 frames; PLM's TVBench from its repository. --: not reported. Gemma~4 reports no video benchmark; MVP, TemporalBench (binary), SSv2 and Vinoground are not reported by any of these models.}
\label{tab:app_reported}
\end{table*}

Pair metrics (MVP, Vinoground) count a pair as correct only if both members are.

\paragraph{Training splits and a train--evaluation audit.} Every source enters training with the training split its authors release; no evaluation split of any benchmark is used. Public collections reuse source videos, so we also matched video identifiers (normalised file names; YouTube identifiers without time suffixes) between all 175 training sources and all 35,488 evaluation items (\cref{tab:app_overlap}). Nine of the twelve sets share no video with training. MVBench and TVBench reuse Charades clips through STAR that LLaVA-Video-178K also contains, and from which one counterfactual source draws clips; TVBench also reuses PerceptionTest training videos; TemporalBench shares YouTube videos with Kinetics-700. In total 147 items (0.41\%) are affected, and removing them changes every model by at most 0.42 points on a benchmark and 0.04 on the twelve-score average, ours and the open models alike. Video LLM reports rarely give such an item-level account of their training data; we report ours in full, and the released manifests list every training clip.
\begin{table}[!htbp]
\centering\footnotesize\setlength{\tabcolsep}{4pt}
\begin{tabular}{lrrl}
\toprule
Benchmark & Items & Shared & Shared source videos \\
\midrule
MVBench & 4,000 & 60 (1.5\%) & Charades (STAR) clips also in LLaVA-Video-178K \\
TVBench & 2,525 & 57 (2.3\%) & the same Charades clips; PerceptionTest train \\
TemporalBench & 2,765 & 30 (1.1\%) & YouTube videos also in Kinetics-700 \\
The other nine sets & 26,198 & 0 & -- \\
\midrule
All twelve & 35,488 & 147 (0.41\%) & \\
\bottomrule
\end{tabular}
\caption{Video-level audit of all 175 training sources against every evaluation item. Rescoring without the shared items moves each model by at most 0.42 points on a benchmark and 0.04 on the twelve-score average, for ours and the open models alike.}
\label{tab:app_overlap}
\end{table}

\begin{table*}[!htbp]
\centering\footnotesize\setlength{\tabcolsep}{3.6pt}
\begin{tabular}{lrrrrrrrrrrrrr}
\toprule
Model & MVB & PT & TVB & TC & TOM & MoB & NQA & MVP & TB & SSv2 & VG-T & VG-V & Avg \\
\midrule
Gemma 4 E2B & 39.8 & 46.3 & 36.4 & 51.5 & 18.7 & 41.1 & 51.4 & 8.1 & 61.4 & 37.0 & 24.0 & 22.4 & 36.51 \\
Gemma 4 E4B & 46.5 & 52.0 & 41.5 & 57.4 & 22.8 & 49.4 & 64.7 & 16.5 & 61.0 & 43.8 & 26.4 & 24.0 & 42.17 \\
\bottomrule
\end{tabular}
\caption{Gemma~4 under the common protocol (thinking off, its own processor and chat template). Its pair-accuracy scores fall far below those of the other open models of its size (MVP 8.1 and 16.5), so it is listed here rather than in \cref{tab:main}; it would rank last in its class either way.}
\label{tab:app_gemma}
\end{table*}

TemporalBench uses the binary short-QA split.
SSv2 is posed as multiple choice over action labels: 12 validation clips for each of the 174 action templates (2,088 items), with four options, the true template and the three most similar other templates by word overlap (\eg ``from left to right'' against ``from right to left''), so the options differ mainly in the action or its direction; the correct letter is spread about evenly over A--D.
Visual tokens per 32-frame clip, measured with each model's processor on 21 benchmark clips from 270p to 1080p: ours 1,024; Gemma~4 2,048--2,112; Molmo2 2,592; VideoLLaMA3 1,006--3,840 (mean 3,168); PLM and InternVL3.5 8,192; Qwen3-VL 1,920--11,664 (mean 7,801), growing with resolution.

\paragraph{Held-out sets.} \Cref{tab:app_heldout_def} defines the six sets. Each group is one source clip under all its variants, so a set tests one novelty with clip content held fixed. Every chosen source video (by file name; for simulation, the whole run) is absent from every training manifest and every evaluation manifest, and the six sets use disjoint clips.
\begin{table}[!htbp]
\centering\footnotesize\setlength{\tabcolsep}{3pt}
\begin{tabular}{@{}lrl>{\raggedright\arraybackslash}p{7.6cm}@{}}
\toprule
Set & Items & Source & What is new \\
\midrule
Speed ratio & 540 & SSv2 validation & relative speed of a clip stacked over itself at span ratios 1.6, 2.5, 4.0 (training: equal, 2:1 and 3:1, and 1.0--1.4 in the fine-motion source; none of the three ratios occurs), plus the equal item of each clip \\
Domain, speed & 600 & Kinetics-700 validation & relative speed (2:1, 3:1, equal) and acceleration on a corpus never used by the speed family \\
Domain, direction & 540 & Kinetics-700 validation & camera direction labeled by optical flow, original / reversed / mirrored, on a corpus unused for this family \\
Combo, direction & 584 & SSv2 validation & mirror \emph{and} reversal together (the answer is the original direction), with the three trained variants of the same clip \\
Combo, speed & 540 & SSv2 validation & acceleration played backwards (accelerating becomes decelerating), plus freeze \\
Camera, domain & 490 & NVIDIA PhysX simulation & synthetic camera moves on a simulator never used for camera rows \\
\bottomrule
\end{tabular}
\caption{Held-out generalization sets (chance 33.3 on average). Question wordings, option pools and option-length balancing are the training builders' own.}
\label{tab:app_heldout_def}
\end{table}
\begin{table}[!htbp]
\centering\footnotesize\setlength{\tabcolsep}{4pt}
{\begin{tabular}{lrrrrrrr}
\toprule
Model & ratio & dom.\,spd & dom.\,dir & comb.\,dir & comb.\,spd & camera & avg \\
\midrule
\multicolumn{8}{@{}l}{\emph{Open models}} \\
Qwen3-VL-2B & 37.0 & 45.5 & 35.2 & 56.5 & 44.6 & 43.9 & 43.8 \\
Qwen3-VL-4B & 35.9 & 50.5 & 40.7 & 73.3 & 50.2 & 49.4 & 50.0 \\
InternVL3.5-4B & 33.9 & 46.7 & 30.7 & 61.1 & 53.0 & 45.1 & 45.1 \\
Molmo2-4B & 38.3 & 50.5 & 29.1 & 59.4 & 51.1 & 32.7 & 43.5 \\
PLM-3B & 35.9 & 49.8 & 35.7 & 64.9 & 58.5 & 44.9 & 48.3 \\
Qwen3-VL-8B & 36.7 & 50.8 & 41.9 & 78.4 & 68.0 & 57.1 & 55.5 \\
InternVL3.5-8B & 33.0 & 45.5 & 33.9 & 63.0 & 52.4 & 49.2 & 46.2 \\
Molmo2-8B & 39.3 & 52.2 & 33.9 & 69.2 & 59.6 & 37.6 & 48.6 \\
PLM-8B & 38.3 & 49.0 & 35.4 & 69.0 & 53.5 & 51.0 & 49.4 \\
VideoLLaMA3-7B & 32.2 & 45.0 & 28.7 & 63.5 & 50.0 & 41.4 & 43.5 \\
\midrule
\multicolumn{8}{@{}l}{\emph{Controlled 2B arms (\cref{tab:2x2}), at 8k steps / final}} \\
SigLIP caps+QA, 8k & 35.0 & 50.0 & 43.0 & 58.0 & 54.3 & 42.2 & 47.1 \\
SigLIP caps+QA, final & 35.9 & 51.0 & 50.4 & 73.6 & 60.0 & 52.9 & 54.0 \\
SigLIP +CF, 8k & 34.8 & 59.8 & 52.0 & 66.3 & 86.3 & 56.5 & 59.3 \\
SigLIP +CF, final & 55.7 & 72.0 & 61.5 & 88.9 & 94.6 & 75.9 & 74.8 \\
V-JEPA caps+QA, 8k & 33.5 & 44.8 & 55.0 & 94.0 & 48.7 & 75.9 & 58.6 \\
V-JEPA caps+QA, final & 33.3 & 44.3 & 69.8 & 97.1 & 49.1 & 92.4 & 64.3 \\
V-JEPA AoT only, 8k & 31.9 & 42.8 & 60.4 & 96.1 & 54.1 & 79.6 & 60.8 \\
V-JEPA AoT only, final & 34.6 & 45.3 & 72.6 & 97.4 & 48.7 & 90.4 & 64.8 \\
V-JEPA temporal QA in place of CF, 8k & 36.1 & 49.2 & 62.2 & 92.8 & 55.0 & 77.6 & 62.1 \\
\rowcolor{ours}V-JEPA +CF, 8k & 91.5 & 93.0 & 68.5 & 97.9 & 97.8 & 99.2 & 91.3 \\
\rowcolor{ours}V-JEPA +CF, final (= Ours-2B-Q) & 98.7 & 98.5 & 75.7 & 98.5 & 98.9 & 99.8 & 95.0 \\
\midrule
\multicolumn{8}{@{}l}{\emph{Ours (final checkpoints)}} \\
\rowcolor{ours}Ours-2B-Q & 98.7 & 98.5 & 75.7 & 98.5 & 98.9 & 99.8 & 95.0 \\
\rowcolor{ours}Ours-1B-L & 97.8 & 99.2 & 71.5 & 98.6 & 99.6 & 100.0 & 94.5 \\
\rowcolor{ours}Ours-4B-Q & 98.9 & 99.0 & 80.9 & 99.0 & 100.0 & 100.0 & 96.3 \\
\rowcolor{ours}Ours-3B-L & 99.1 & 99.2 & 78.9 & 99.0 & 100.0 & 100.0 & 96.0 \\
\rowcolor{ours}Ours-8B-Q & 99.1 & 99.0 & 78.5 & 99.1 & 99.4 & 100.0 & 95.9 \\
\rowcolor{ours}Ours-8B-L & 99.3 & 98.7 & 80.0 & 98.8 & 99.6 & 100.0 & 96.1 \\
\bottomrule
\end{tabular}}
\caption{Held-out construction sets (definitions in \cref{tab:app_heldout_def}: unseen speed ratios, an unseen source for the speed and direction families, unseen combinations, a simulator unused for camera rows). Ratio: unseen playback-speed ratios (chance 33.3); dom.: speed / direction on a new source corpus; comb.: unseen mirror\,$\times$\,reversal and acceleration\,$\times$\,reversal combinations; camera: simulated camera motion. 4B-class open models average 43.5--50.0 and Qwen3-VL-2B 43.8.}
\label{tab:app_heldout}
\end{table}

\begin{table}[!htbp]
\centering\footnotesize\setlength{\tabcolsep}{4pt}
\begin{tabular}{lrrr}
\toprule
& \multicolumn{2}{c}{inside training range} & beyond \\
\cmidrule(lr){2-3}\cmidrule(l){4-4}
Model & 1.6 & 2.5 & 4.0 \\
\midrule
\multicolumn{4}{@{}l}{\emph{Controlled 2B arms, 8k}} \\
SigLIP +CF & 38.3 & 35.6 & 30.6 \\
V-JEPA caps+QA & 32.8 & 34.4 & 33.3 \\
V-JEPA temporal QA in place of CF & 37.8 & 36.7 & 33.9 \\
\rowcolor{ours}V-JEPA +CF & 88.9 & 91.7 & 93.9 \\
\midrule
\multicolumn{4}{@{}l}{\emph{Ours (final checkpoints)}} \\
\rowcolor{ours}Ours-2B-Q & 97.2 & 99.4 & 99.4 \\
\rowcolor{ours}Ours-1B-L & 94.4 & 99.4 & 99.4 \\
\rowcolor{ours}Ours-4B-Q & 97.2 & 100.0 & 99.4 \\
\rowcolor{ours}Ours-3B-L & 97.8 & 99.4 & 100.0 \\
\rowcolor{ours}Ours-8B-Q & 97.8 & 99.4 & 100.0 \\
\rowcolor{ours}Ours-8B-L & 97.8 & 100.0 & 100.0 \\
\bottomrule
\end{tabular}
\caption{Unseen speed ratios separately (180 items each, chance 33.3). Training uses equal, 2:1 and 3:1 and fine ratios up to 1.4, so 1.6 and 2.5 lie between trained ratios and 4.0 beyond them.}
\label{tab:app_ratio}
\end{table}

\paragraph{Successive playback.} Training rows convey a relative speed by stacking two playbacks of a clip and an acceleration by a continuous time warp. \Cref{tab:app_seq} keeps the 180 clips and unseen ratios (1.6, 2.5, 4.0) of the speed-ratio set but changes the construction: the clip is played twice one after the other in one 32-frame input, its short central span and then its full span, or the reverse, so the motion speeds up or slows down halfway through. No training row builds speed this way. The question and options are the acceleration family's (``How does the motion change over the video?''; speeding up, slowing down, not moving). Our six models reach 92.5--99.2 (pair 85.0--98.3) against 19.4--51.9 for open models, whose pair accuracy of 0.6--21.7 (chance 11.1) shows that they mostly give both orders of a clip the same answer; without CF the V-JEPA arm stays at 39.4.
\begin{table}[!htbp]
\centering\footnotesize\setlength{\tabcolsep}{6pt}
\begin{tabular}{lrr}
\toprule
Model & Single & Pair \\
\midrule
\multicolumn{3}{@{}l}{\emph{Open models}} \\
Qwen3-VL-2B & 33.1 & 8.9 \\
Qwen3-VL-4B & 36.9 & 1.1 \\
InternVL3.5-4B & 24.2 & 0.6 \\
Molmo2-4B & 19.4 & 4.4 \\
PLM-3B & 29.2 & 2.8 \\
Qwen3-VL-8B & 51.9 & 21.7 \\
InternVL3.5-8B & 22.5 & 1.7 \\
Molmo2-8B & 43.1 & 8.9 \\
PLM-8B & 49.2 & 16.7 \\
VideoLLaMA3-7B & 35.0 & 10.6 \\
\midrule
\multicolumn{3}{@{}l}{\emph{Controlled 2B arms (\cref{tab:2x2})}} \\
V-JEPA caps+QA, final & 39.4 & 6.7 \\
\midrule
\multicolumn{3}{@{}l}{\emph{Ours (final checkpoints)}} \\
\rowcolor{ours}Ours-2B-Q & 93.9 & 87.8 \\
\rowcolor{ours}Ours-1B-L & 92.5 & 85.0 \\
\rowcolor{ours}Ours-4B-Q & 96.7 & 93.3 \\
\rowcolor{ours}Ours-3B-L & 95.8 & 91.7 \\
\rowcolor{ours}Ours-8B-Q & 93.9 & 87.8 \\
\rowcolor{ours}Ours-8B-L & 99.2 & 98.3 \\
\bottomrule
\end{tabular}
\caption{Successive playback (360 items, 180 clips): the speed-ratio set's clips and unseen ratios, the clip played twice one after the other, asked whether the motion speeds up or slows down. Single: chance 33.3; pair: both orders of a clip correct, chance 11.1.}
\label{tab:app_seq}
\end{table}

\paragraph{Single-frame control.} With the middle frame repeated (no motion, layout and composition unchanged), the six held-out sets drop from 95.0 to 37.9 (2B) and from 96.3 to 37.6 (4B), against a chance level of 33.3: layout and single-frame cues do not solve them.

\paragraph{Inference cost.} \Cref{tab:app_speed} separates the vision encoder, the connector and the language model; all models are timed and their FLOPs counted by the same script on the same two benchmark clips (MVBench, 480p; TVBench, 720p). The two parts behave differently. The encoder's cost is a property of the chosen encoder: V-JEPA~2.1 attends jointly over all 9,216 tubelet patches of a 32-frame clip (attention is 60\% of its 13.9 TFLOPs for ViT-L and 51\% of 37.8 for ViT-g), and it is the same for every recipe on that encoder and every language-model size. The open models use frame-wise image encoders (SigLIP-2-based in Qwen3-VL~\cite{qwen3vl,tschannen2025siglip2}, InternViT in InternVL3.5, PE in PLM) that attend within one frame or frame pair at a time, so their encoder cost grows linearly with the number of frames and no attention crosses time. Qwen3-VL-2B's encoder has the depth and width of our ViT-L (24 layers, width 1,024) and takes 35\,ms against 60\,ms at 480p; the difference is V-JEPA's joint space-time attention, the property that carries the dynamics (\cref{sec:measure}). This alone orders the 480p first token in the 2B class (Qwen3-VL-2B 66\,ms, Ours-1B-L 70, Ours-2B-Q 83); at 720p Qwen3-VL's token count grows and the order reverses (308 against 71 and 82). The language model's cost is set by the number of visual tokens and grows with its size; this is the part our design reduces, since the bridge passes 1,024 tokens per clip at any resolution. Per visual token, prefill costs 2.1--14.6 GFLOPs from 1B to 8B, so the saving over models with 8,192 tokens grows with scale: 22--37 TFLOPs at 2B, 62--96 at 4B and 136--145 at 8B, each more than our encoder's 14 or 38.
\begin{table*}[!htbp]
\centering\footnotesize\setlength{\tabcolsep}{4pt}
\begin{tabular}{@{}lrrrrrrrrrr@{}}
\toprule
 & & \multicolumn{2}{c}{vision encoder} & connector & \multicolumn{2}{c}{LLM prefill} & first token & decode & \multicolumn{2}{c}{720p clip (ms)} \\ \cmidrule(lr){3-4}\cmidrule(lr){6-7}\cmidrule(l){10-11}
Model & vis.\ tokens & TFLOPs & ms & ms & TFLOPs & ms & ms & ms/token & LLM prefill & first token \\
\midrule
\multicolumn{11}{@{}l}{\emph{2B class}} \\
Qwen3-VL-2B & 1,920 / 11,520 & 5.0 & 35 & in enc. & 7.0 & 31 & 66 & 18.2 & 154 & 308 \\
InternVL3.5-2B & 8,192 & 23.2 & 55 & 0.4 & 40.5 & 104 & 159 & 18.4 & 104 & 159 \\
PLM-1B & 8,192 & 22.2 & 87 & 1.8 & 25.2 & 57 & 146 & 9.3 & 57 & 146 \\
\rowcolor{ours}Ours-2B-Q & 1,024 & 13.9 & 60 & 0.8 & 3.3 & 22 & 83 & 20.0 & 21 & 82 \\
\rowcolor{ours}Ours-1B-L & 1,024 & 13.9 & 60 & 0.6 & 2.3 & 10 & 70 & 8.2 & 10 & 71 \\
\multicolumn{11}{@{}l}{\emph{3B--4B class}} \\
Qwen3-VL-4B & 1,920 / 11,520 & 5.0 & 35 & in enc. & 18.1 & 58 & 93 & 24.9 & 319 & 474 \\
InternVL3.5-4B & 8,192 & 23.2 & 55 & 0.5 & 104.4 & 218 & 273 & 25.3 & 218 & 273 \\
Molmo2-4B & 2,592 & 19.7 & 132 & in enc. & 26.1 & 93 & 225 & 27.8 & 103 & 235 \\
PLM-3B & 8,192 & 22.2 & 87 & 3.0 & 70.4 & 138 & 228 & 15.6 & 138 & 228 \\
\rowcolor{ours}Ours-4B-Q & 1,024 & 37.8 & 137 & 0.9 & 8.6 & 29 & 167 & 23.9 & 29 & 168 \\
\rowcolor{ours}Ours-3B-L & 1,024 & 37.8 & 138 & 0.9 & 6.7 & 20 & 159 & 15.1 & 20 & 159 \\
\multicolumn{11}{@{}l}{\emph{8B class}} \\
Qwen3-VL-8B & 1,920 / 11,520 & 6.8 & 43 & in enc. & 32.2 & 81 & 124 & 25.2 & 450 & 666 \\
InternVL3.5-8B & 8,192 & 23.2 & 55 & 0.8 & 160.7 & 314 & 370 & 24.5 & 313 & 369 \\
Molmo2-8B & 2,592 & 19.7 & 133 & in enc. & 45.4 & 127 & 261 & 28.4 & 137 & 271 \\
PLM-8B & 8,192 & 123.6 & 336 & 4.6 & 151.9 & 270 & 610 & 18.3 & 271 & 610 \\
\rowcolor{ours}Ours-8B-Q & 1,024 & 37.8 & 139 & 0.9 & 15.8 & 42 & 182 & 24.4 & 42 & 182 \\
\rowcolor{ours}Ours-8B-L & 1,024 & 37.8 & 138 & 0.9 & 16.2 & 36 & 175 & 17.1 & 36 & 175 \\
\bottomrule
\end{tabular}
\caption{Inference cost per 32-frame clip on one H100 (batch 1, bf16, video decoding excluded; times: median of 10 runs after 3 warm-ups; 480p MVBench clip unless noted). TFLOPs: matrix multiplications including attention, counted with PyTorch's FLOP counter. Vision encoder: the vision tower; for our models the frozen, unmodified V-JEPA~2.1 forward (ViT-L for 1B and 2B, ViT-g otherwise), a cost set by the choice of encoder and identical for any recipe built on it: V-JEPA attends jointly over all frames, while the open models' encoders attend within a frame or frame pair. Connector: projector or bridge (``in enc.'': part of the vision module; at most 1.5 TFLOPs for every model). LLM prefill: the language model over the whole prompt; first token: all three parts; decode: per generated token with KV cache (128 tokens). Visual tokens: 480p / 720p where they differ (Molmo2: from its processor). Our token count and cost do not depend on resolution; Qwen3-VL's grow with it. --: not timed.}
\label{tab:app_speed}
\end{table*}

\FloatBarrier
\section{Encoder measurements}
\label{app:enc}
\begin{table}[!htbp]
\centering\footnotesize\setlength{\tabcolsep}{2.6pt}
{\begin{tabular}{llrrrrr}
\toprule
Encoder & Transf. & $r_d$ & $\kappa$ & $\rho$ & proto 8 & proto all \\
\midrule
SigLIP so400m & reversal & 0 & -- & -- & 50.0 & 50.0 \\
 & speed & .016 & .0016 & .18 & 50.0 & 50.3 \\
 & mirror & .066 & .0014 & .18 & 49.8 & 49.0 \\
V-JEPA ViT-L & reversal & .064 & .037 & .12 & 58.8 & 63.7 \\
 & speed & .055 & .080 & .40 & 66.3 & 71.0 \\
 & mirror & .093 & .031 & .031 & 58.7 & 70.4 \\
V-JEPA ViT-g & reversal & .108 & .039 & .075 & 63.9 & 74.3 \\
 & speed & .080 & .167 & .59 & 77.2 & 78.8 \\
 & mirror & .104 & .074 & .029 & 69.2 & 81.2 \\
\midrule
\multicolumn{7}{@{}l}{\emph{Other video encoders (diagnostic)}} \\
ViCLIP ViT-L/14 & reversal & .052 & .0012 & .24 & 49.8 & 49.7 \\
 & speed & .057 & .042 & .17 & 60.1 & 71.2 \\
 & mirror & .029 & .0019 & .14 & 50.2 & 50.3 \\
VideoMAE ViT-L & reversal & .502 & .0032 & .41 & 50.6 & 55.8 \\
 & speed & .370 & .632 & .17 & 100.0 & 100.0 \\
 & mirror & .172 & .045 & .094 & 64.7 & 79.3 \\
\bottomrule
\end{tabular}}
\caption{Encoder decomposition on 1,200 clips (token-averaged $d$; 70\% fit, 30\% held out). $\kappa$ uses $y=+1$ for every reversal and speed pair (the direction of the transformation) and $y=\pm1$ for left-to-right / right-to-left motion under mirroring, on frozen features pooled to $4{\times}4$ cells per tubelet and standardised per dimension on the fitting clips; random directions give $\kappa\approx1/840\approx0.001$. $\rho$: overlap of the change direction with the 32 leading content components (a content direction scores 0.44--0.63). SigLIP's reversal component is zero by construction. ViCLIP~\cite{wang2024internvid}: video--text contrastive training with joint attention over all frame tokens (16 frames, the 8-frame temporal embedding interpolated); VideoMAE~\cite{tong2022videomae}: masked pixel prediction with $2\times16\times16$ tubelets (32 frames); both at $224^2$ (ViCLIP: penultimate block; VideoMAE: last layer), on the same clips, variants and pooling.}
\label{tab:app_enc}
\end{table}

\Cref{tab:app_map} lines up, for each family, the transformation, the theoretical prediction, the measured $\kappa$ and the gain from counterfactual supervision on the matching ability.
\begin{table*}[!htbp]
\centering\footnotesize\setlength{\tabcolsep}{4pt}
\resizebox{\textwidth}{!}{\begin{tabular}{llllrrrrl}
\toprule
& & & & \multicolumn{2}{c}{$\kappa$ (frozen)} & \multicolumn{2}{c}{CF gain at 8k} & \\
\cmidrule(lr){5-6}\cmidrule(lr){7-8}
Family & Transformation & Theory & Prediction & V-JEPA & SigLIP & V-JEPA & SigLIP & Ability \\
\midrule
Speed, rate & rescaling $\tau_a$ & \cref{thm:b}(i); measured $\kappa$ & large on V-JEPA, small on SigLIP & .080 & .0016 & $+32.8$ & $+11.4$ & speed \\
Sign flips & reversal, camera moves & \cref{thm:a}: $d\in V_-$; \cref{thm:b}(i): $d\equiv0$ & large on V-JEPA & .037 & -- & $+6.8$ & $0.0$ & camera \\
 & & & & & & $+4.1$ & $-0.4$ & state change \\
Sign flips & mirror & \cref{thm:a}; \cref{thm:b}(ii) & large on V-JEPA & .031 & .0014 & $+3.7$ & $+0.1$ & direction \\
Event order & segment permutation & local receptive field: $A=I$ & equal on both & -- & -- & $+1.9$ & $+1.7$ & order \\
\bottomrule
\end{tabular}}
\caption{From encoder structure to gain. For each family: the transformation, the result that predicts how its change is carried, the prediction, the measured cross-video consistency $\kappa$ of the frozen ViT-L and SigLIP (\cref{tab:app_enc}; SigLIP reversal has $r_d=0$, no in-token component), and the CF gain on the matching natural-benchmark ability at 8k steps (\cref{tab:app_skills}). Order changes token positions rather than token content, so both encoders gain alike. Speed is the one family the propositions do not cover for V-JEPA (a rate change gives no shared direction by \cref{thm:a}, \cref{app:theory}); its prediction rests on the measured $\kappa$ alone, with \cref{thm:b}(i) giving SigLIP only scene-specific frame differences. Gains are those of the whole counterfactual mixture against captions and QA, read per ability; at the final step the direction and state-change gains of the two encoders draw closer, while speed stays far apart (\cref{tab:app_skills}).}
\label{tab:app_map}
\end{table*}

\paragraph{Linear probes} on frozen features (time direction / speed / motion direction) give SigLIP 65.3 / 73.5 / 97.4, ViCLIP 56.9 / 70.3 / 97.5, VideoMAE 65.1 / 76.8 / 95.7 and V-JEPA ViT-L 80.8 / 81.2 / 98.8: V-JEPA reads time direction and speed best. These are two-class probes on the clip pairs of \cref{sec:measure} (a clip against its reversal, its half-speed copy or its mirror image); the speed-ratio items of \cref{tab:unused} are three-way between ratios, where a frozen SigLIP probe reads 36.4 and V-JEPA 86.7.
\emph{Prototype readout} (proto 8 / all): the readout $w$ is the mean of $y\,\bar d$ over 8 (or all) fitting pairs, with $\bar d$ the token-averaged change, the first gradient step of \cref{prop:kappa}; it is applied to single held-out clips (token-averaged $z$, threshold at the fitting clips' mean content), so it tests whether a direction read from few pairs classifies new videos; this includes the content term of \cref{prop:kappa}, which $\rho$ bounds.
ViT-g gives 79.7 / 84.7 / 98.6.
Motion direction is solvable from the first time step's position alone for both encoders, so it does not separate them.

\FloatBarrier
\section{Encoder $\times$ supervision in full}
\label{app:2x2}
\Cref{tab:app_2x2} lists every checkpoint and benchmark of the four 2B arms.
\Cref{tab:app_pool} pools four checkpoints per comparison to average out checkpoint noise.
\Cref{tab:app_skills} lists all fourteen ability groups. The speed group combines two benchmarks, and the gain holds on each: at the final step CF adds 62.9 points on TOMATO's speed items (210; $[+56.2,+69.5]$) and 21.8 on TempCompass's (317; $[+14.8,+28.7]$) for V-JEPA, against 41.9 and 0.3 for SigLIP.
\emph{4B control.} The 4B model trained without the counterfactual rows, otherwise identical (mixture, schedule, global batch 150), to 8k: twelve-score average 61.28 against 66.06 with them ($+4.8$ $[+4.0,+5.5]$, paired bootstrap), speed items $+29.4$ $[+24.5,+34.2]$, held-out sets 61.4 against 93.0 (unseen ratios 35.7 against 93.3), and successive playback (\cref{tab:app_seq}) 33.1 against 90.0 (pair 10.6 against 80.0).

\emph{AoT-only arm.} The 33 counterfactual sources are replaced, at their total weight, by one source of 60k rows: 30k SSv2 training clips from the pool the counterfactual sources already use, each forward and reversed, with ``Is this video playing forward or in reverse?'' or one of four rewordings and two shuffled options (answers balanced); everything else is the V-JEPA arm's recipe, trained to 8k and then to the final step like the caps+QA arm. On 2,088 unseen SSv2 validation clips, each forward and reversed, it answers this question correctly on 95.9\% of videos and on both videos of 93.3\% of clips at the final step (94.4 and 90.5 at 8k; caps+QA: 50.0), so it learns its target; the benchmarks do not improve ($-0.4$ $[-1.0,+0.2]$ against caps+QA at the final step, paired bootstrap; \cref{tab:2x2}), and speed items stay at the caps+QA level (34.0 against 33.4 at the end).
\begin{table*}[!htbp]
\centering\footnotesize\setlength{\tabcolsep}{4pt}
\begin{tabular}{lrrrrrrrrrrrrr}
\toprule
Step & MVB & PT & TVB & TC & TOM & MoB & NQA & MVP & TB & SSv2 & VG-T & VG-V & Avg \\
\midrule
\multicolumn{14}{l}{\textit{SigLIP caps+QA}} \\
2k & 51.6 & 59.2 & 49.2 & 59.6 & 23.6 & 44.0 & 66.8 & 12.8 & 52.2 & 67.3 & 34.8 & 23.2 & 45.36 \\
4k & 55.5 & 62.8 & 50.4 & 62.6 & 25.9 & 46.3 & 68.6 & 30.4 & 54.6 & 71.6 & 48.6 & 23.6 & 50.07 \\
6k & 59.9 & 65.5 & 56.7 & 63.3 & 26.9 & 47.5 & 70.6 & 36.1 & 55.2 & 73.0 & 50.0 & 25.0 & 52.48 \\
8k & 62.7 & 65.8 & 58.8 & 63.1 & 26.9 & 48.1 & 71.7 & 37.7 & 58.1 & 74.8 & 49.0 & 24.8 & 53.46 \\
12k & 64.5 & 68.2 & 60.1 & 65.3 & 29.5 & 48.7 & 72.8 & 44.9 & 57.9 & 79.5 & 50.6 & 25.6 & 55.63 \\
16k & 66.4 & 69.7 & 62.2 & 64.9 & 30.3 & 48.6 & 74.1 & 47.6 & 58.6 & 80.0 & 53.6 & 25.4 & 56.78 \\
20k & 66.5 & 71.4 & 64.1 & 65.5 & 32.3 & 49.5 & 75.1 & 47.3 & 61.4 & 84.6 & 50.8 & 24.2 & 57.72 \\
24k & 67.8 & 72.8 & 63.2 & 65.1 & 30.9 & 50.8 & 75.6 & 48.3 & 60.7 & 84.9 & 51.6 & 25.6 & 58.11 \\
final & 67.7 & 72.5 & 64.4 & 65.6 & 33.0 & 50.8 & 75.6 & 48.2 & 61.6 & 84.5 & 51.8 & 26.6 & 58.52 \\
\midrule
\multicolumn{14}{l}{\textit{SigLIP +CF}} \\
2k & 50.2 & 59.6 & 47.8 & 60.6 & 27.4 & 42.1 & 65.3 & 18.9 & 47.1 & 68.1 & 44.2 & 24.6 & 46.32 \\
4k & 55.4 & 62.9 & 55.8 & 63.3 & 26.2 & 44.1 & 70.1 & 29.1 & 54.7 & 69.3 & 44.6 & 33.6 & 50.76 \\
6k & 57.5 & 64.4 & 56.6 & 65.1 & 34.9 & 44.5 & 70.7 & 37.8 & 57.4 & 73.1 & 48.6 & 34.8 & 53.78 \\
8k & 61.5 & 66.6 & 58.8 & 66.3 & 32.3 & 46.5 & 72.4 & 42.7 & 58.6 & 78.6 & 57.0 & 40.0 & 56.77 \\
12k & 63.2 & 67.7 & 59.4 & 65.9 & 33.5 & 48.1 & 72.8 & 43.5 & 56.7 & 80.4 & 51.2 & 40.0 & 56.87 \\
16k & 66.4 & 70.8 & 63.0 & 67.5 & 37.7 & 50.1 & 74.7 & 50.0 & 61.7 & 83.7 & 54.0 & 42.6 & 60.18 \\
20k & 66.3 & 71.9 & 64.6 & 67.4 & 38.5 & 49.4 & 75.1 & 50.6 & 61.3 & 83.5 & 52.4 & 45.8 & 60.57 \\
24k & 67.7 & 72.4 & 66.2 & 68.8 & 38.9 & 50.7 & 75.7 & 52.2 & 63.0 & 85.3 & 54.0 & 45.8 & 61.73 \\
final & 68.3 & 71.9 & 67.6 & 70.0 & 38.9 & 50.4 & 76.0 & 53.1 & 63.4 & 84.4 & 54.8 & 46.8 & 62.13 \\
\midrule
\multicolumn{14}{l}{\textit{V-JEPA caps+QA}} \\
2k & 53.9 & 59.6 & 53.3 & 57.0 & 26.6 & 43.9 & 63.8 & 25.0 & 54.0 & 78.5 & 38.0 & 23.4 & 48.08 \\
4k & 59.1 & 64.6 & 59.0 & 62.5 & 27.8 & 48.7 & 67.8 & 35.0 & 58.6 & 82.6 & 43.8 & 23.4 & 52.74 \\
6k & 62.1 & 67.6 & 63.3 & 63.4 & 30.5 & 49.0 & 68.9 & 40.5 & 58.1 & 83.5 & 51.4 & 25.8 & 55.34 \\
8k & 64.2 & 68.2 & 67.6 & 63.2 & 31.7 & 50.2 & 70.3 & 42.1 & 58.8 & 86.3 & 54.6 & 22.6 & 56.65 \\
12k & 66.5 & 70.3 & 65.1 & 65.1 & 35.2 & 51.9 & 71.6 & 46.6 & 60.0 & 88.4 & 54.6 & 25.6 & 58.41 \\
16k & 68.2 & 71.4 & 66.7 & 65.9 & 37.8 & 53.1 & 72.9 & 48.5 & 61.1 & 89.0 & 60.4 & 23.0 & 59.83 \\
20k & 68.9 & 72.9 & 70.2 & 67.5 & 37.5 & 53.4 & 73.5 & 49.9 & 62.5 & 90.3 & 58.8 & 25.0 & 60.87 \\
24k & 69.3 & 73.4 & 70.7 & 66.7 & 38.4 & 54.7 & 74.2 & 50.3 & 62.9 & 90.7 & 58.8 & 25.2 & 61.27 \\
final & 69.9 & 74.2 & 71.7 & 66.5 & 37.5 & 54.1 & 74.4 & 51.7 & 63.7 & 90.2 & 58.4 & 26.0 & 61.53 \\
\midrule
\multicolumn{14}{l}{\textit{V-JEPA AoT only}} \\
8k & 63.5 & 68.2 & 62.8 & 63.5 & 29.6 & 49.4 & 69.4 & 40.6 & 61.6 & 83.6 & 58.8 & 25.4 & 56.37 \\
final & 69.6 & 73.6 & 69.0 & 66.1 & 36.8 & 53.8 & 73.8 & 50.9 & 65.5 & 89.1 & 62.0 & 23.8 & 61.17 \\
\midrule
\multicolumn{14}{l}{\textit{V-JEPA temporal QA in place of CF}} \\
8k & 65.9 & 69.1 & 66.5 & 64.5 & 37.0 & 53.1 & 70.9 & 44.9 & 60.0 & 85.2 & 57.2 & 24.2 & 58.21 \\
\midrule
\multicolumn{14}{l}{\textit{V-JEPA +CF}} \\
\rowcolor{ours}2k & 49.5 & 60.6 & 53.1 & 61.3 & 34.3 & 44.3 & 60.5 & 30.3 & 54.3 & 75.5 & 50.8 & 29.2 & 50.31 \\
\rowcolor{ours}4k & 58.2 & 65.5 & 63.2 & 64.9 & 38.7 & 45.9 & 66.9 & 39.7 & 56.9 & 79.8 & 53.2 & 37.4 & 55.86 \\
\rowcolor{ours}6k & 59.1 & 66.3 & 60.6 & 67.3 & 45.6 & 48.2 & 68.6 & 42.0 & 59.4 & 81.4 & 56.0 & 46.6 & 58.43 \\
\rowcolor{ours}8k & 63.0 & 69.7 & 63.4 & 69.6 & 48.6 & 49.5 & 69.9 & 45.9 & 61.4 & 86.6 & 61.4 & 45.2 & 61.18 \\
\rowcolor{ours}10k & 64.5 & 70.8 & 64.6 & 69.5 & 49.8 & 52.4 & 70.9 & 47.9 & 63.3 & 87.4 & 60.6 & 48.2 & 62.49 \\
\rowcolor{ours}12k & 64.7 & 70.3 & 64.9 & 69.9 & 48.0 & 51.3 & 71.2 & 46.3 & 59.6 & 88.5 & 60.6 & 37.6 & 61.07 \\
\rowcolor{ours}14k & 67.1 & 72.5 & 67.8 & 71.7 & 48.3 & 54.0 & 71.8 & 51.5 & 62.3 & 89.3 & 61.2 & 38.8 & 63.02 \\
\rowcolor{ours}16k & 68.3 & 73.0 & 67.0 & 73.2 & 52.0 & 52.5 & 72.8 & 51.2 & 62.6 & 89.6 & 63.6 & 48.6 & 64.53 \\
\rowcolor{ours}18k & 68.4 & 73.3 & 67.7 & 71.5 & 49.6 & 53.2 & 73.1 & 53.8 & 62.8 & 89.5 & 64.2 & 47.8 & 64.58 \\
\rowcolor{ours}20k & 68.0 & 74.0 & 67.3 & 73.0 & 50.3 & 53.9 & 73.3 & 55.5 & 64.2 & 90.0 & 62.8 & 48.2 & 65.04 \\
\rowcolor{ours}22k & 69.3 & 74.3 & 70.0 & 73.2 & 49.6 & 54.3 & 73.8 & 54.9 & 64.3 & 90.7 & 66.8 & 48.8 & 65.83 \\
\rowcolor{ours}24k & 69.3 & 75.0 & 69.6 & 74.2 & 50.4 & 53.8 & 74.3 & 54.6 & 63.1 & 91.1 & 63.8 & 44.8 & 65.33 \\
\rowcolor{ours}26k & 69.6 & 75.2 & 70.4 & 73.6 & 49.4 & 54.2 & 74.1 & 54.5 & 64.4 & 91.5 & 64.6 & 47.4 & 65.74 \\
\rowcolor{ours}final & 69.5 & 75.3 & 70.7 & 73.5 & 50.9 & 54.3 & 74.0 & 54.6 & 63.9 & 91.6 & 63.2 & 50.8 & 66.02 \\
\bottomrule
\end{tabular}
\caption{Encoder $\times$ supervision, every evaluated checkpoint (2B, global batch 168). The V-JEPA caps+QA arm was trained to 8k first and then continued to the final step; the AoT-only arm likewise, and the temporal-QA arm was trained to 8k. Temporal QA: the counterfactual rows removed and their weight given to the 22 annotation-based temporal-QA sources of the mixture (temporal grounding, highlights, step order, before/after, next action, repetition counting, unintentional action).}
\label{tab:app_2x2}
\end{table*}

\begin{table*}[!htbp]
\centering\footnotesize\setlength{\tabcolsep}{4pt}
\begin{tabular}{llrrrrrrrrrrr}
\toprule
Comparison & Steps & MVB & PT & TVB & TC & TOM & MoB & NQA & MVP & SSv2 & VG-T & all items \\
\midrule
SigLIP+CF $-$ V-JEPA+CF & 12--24k & -1.7 & -2.4 & -3.9 & -5.2 & -13.0 & -3.3 & +1.7 & -1.6 & -6.5 & -5.7 & -2.37 \\
SigLIP+CF $-$ SigLIP caps & 12--24k & -0.4 & +0.2 & +0.9 & +2.2 & +6.4 & +0.2 & +0.2 & +0.2 & +1.0 & +0.5 & +0.63 \\
SigLIP caps $-$ V-JEPA+CF & 12--24k & -1.3 & -2.6 & -4.8 & -7.4 & -19.4 & -3.5 & +1.5 & -1.8 & -7.6 & -6.2 & -3.00 \\
V-JEPA caps $-$ V-JEPA+CF & 2--8k & +2.4 & -0.5 & +0.7 & -4.3 & -12.6 & +1.0 & +1.2 & -1.6 & +1.9 & -4.7 & -0.22 \\
SigLIP+CF $-$ V-JEPA+CF & 2--8k & -1.3 & -2.1 & -5.3 & -2.0 & -11.6 & -2.7 & +3.1 & -3.6 & -8.5 & -3.9 & -1.97 \\
SigLIP caps $-$ V-JEPA caps & 2--8k & -2.4 & -1.7 & -7.0 & +0.6 & -3.4 & -1.5 & +1.7 & -2.6 & -11.1 & -0.5 & -1.89 \\
SigLIP+CF $-$ SigLIP caps & 2--8k & -1.3 & +0.0 & +1.0 & +1.6 & +4.4 & -2.2 & +0.2 & +0.6 & +0.6 & +1.4 & +0.13 \\
\bottomrule
\end{tabular}
\caption{Paired item-level differences pooled over four checkpoints (each item's mean difference over the steps; bootstrap over items). +CF: with counterfactual supervision; caps: caption and QA mixture alone. 95\% intervals exclude zero for every entry with $|\Delta|\geq1.6$ (e.g.\ TOMATO $-13.0$ $[-15.2,-10.8]$, SSv2 $-6.5$ $[-7.7,-5.4]$, NExT-QA $+1.7$ $[+1.0,+2.3]$ in the first row). ``All items'' weights benchmarks by size (NExT-QA has 8.6k items).}
\label{tab:app_pool}
\end{table*}

\begin{table*}[!htbp]
\centering\footnotesize\setlength{\tabcolsep}{3.2pt}
\begin{tabular}{lrrrrrrrrrrrrrr}
\toprule
Comparison & speed & direction & order & camera & count & state & localiz. & antonym & rotation & tracking & TB & SSv2 & NQA & knowl. \\
\midrule
CF on V-JEPA, 8k & +32.8 & +3.7 & +1.9 & +6.8 & -1.5 & +4.1 & -5.0 & +3.8 & +2.8 & -8.9 & +2.6 & +0.3 & -0.4 & -0.4 \\
CF on V-JEPA, final & +38.1 & +4.0 & +1.9 & +3.6 & +1.6 & +1.5 & -3.1 & -1.0 & +2.1 & -4.0 & +0.2 & +1.3 & -0.4 & +0.0 \\
AoT only on V-JEPA, 8k & +0.6 & -3.9 & -0.4 & +3.4 & -6.5 & -0.4 & -4.7 & +1.2 & +1.7 & -3.3 & +2.8 & -2.6 & -0.9 & -0.5 \\
CF on SigLIP, 8k & +11.4 & +0.1 & +1.7 & +0.0 & +1.2 & -0.4 & -0.6 & +7.1 & +3.8 & -8.0 & +0.5 & +3.9 & +0.7 & -0.8 \\
CF on SigLIP, 24k & +18.0 & +3.0 & +3.3 & +1.9 & +1.4 & -0.3 & +1.7 & +5.4 & -0.3 & +3.8 & +2.4 & +0.4 & +0.1 & -0.3 \\
CF on SigLIP, final & +16.9 & +3.2 & +3.4 & +1.2 & +2.3 & +1.5 & +2.5 & +5.2 & -1.0 & +0.2 & +1.7 & -0.1 & +0.4 & -0.4 \\
V-JEPA $-$ SigLIP, caps, 8k & -1.5 & +19.7 & +3.4 & +4.1 & +7.7 & -2.3 & +5.6 & +0.8 & +5.9 & +0.9 & +0.7 & +11.5 & -1.4 & +0.6 \\
V-JEPA $-$ SigLIP, +CF, 8k & +19.9 & +23.3 & +3.6 & +10.9 & +5.0 & +2.3 & +1.1 & -2.5 & +4.9 & +0.0 & +2.8 & +8.0 & -2.4 & +1.1 \\
V-JEPA $-$ SigLIP, +CF, 24k & +15.7 & +17.9 & +2.5 & +8.5 & +7.9 & +3.4 & +2.5 & -0.6 & -1.0 & +0.0 & +0.1 & +5.7 & -1.4 & +0.8 \\
\bottomrule
\end{tabular}
\caption{Ability groups over the natural benchmarks' own categories (2B; final: step 27,382). Items: speed 527, direction 1,170, order 1,457, camera 585, count 1,776, state change 711, localization 360, reversal/antonym 520, rotation 286, tracking 425, TemporalBench 2,765, SSv2 2,088, NExT-QA 8,564, knowledge (objects, scenes, knowledge) 9,254. Paired bootstrap 95\% intervals: speed on V-JEPA $[+27.9,+37.8]$, on SigLIP $[+6.3,+16.5]$ (8k) and $[+12.7,+23.3]$ (24k); direction between encoders without CF $[+16.7,+22.7]$; speed between encoders without CF $[-4.9,+1.7]$.}
\label{tab:app_skills}
\end{table*}

\paragraph{Speed in the caption and QA mixture.} Without the counterfactual rows the mixture is not silent about speed: in the temporal-QA arm's mixture, 1.6\% of drawn rows contain \emph{faster}, \emph{slower}, \emph{quicker}, \emph{speed}, \emph{slow motion}, \emph{time-lapse} or \emph{accelerat-}/\emph{decelerat-} (mostly LLaVA-Video captions and SSv2 descriptions), about 21k rows over the first 8k steps. The answer still stays at chance on unseen speed ratios (36.1, \cref{tab:app_heldout}): the words are there, but nothing ties the answer to the change.

\FloatBarrier
\section{Counterfactual consistency in full}
\label{app:heldout}
\begin{table*}[!htbp]
\centering\footnotesize\setlength{\tabcolsep}{5pt}
\begin{tabular}{lrrrrrrrr}
\toprule
& \multicolumn{4}{c}{reversal (1,350 items, 5 benchmarks)} & \multicolumn{4}{c}{mirror (811 items)} \\
\cmidrule(lr){2-5}\cmidrule(lr){6-9}
Model & orig & follows & both & stuck & orig & follows & both & stuck \\
\midrule
Molmo2-4B & 59.9 & 51.5 & 31.9 & 41.6 & 48.7 & 44.6 & 23.9 & 45.5 \\
InternVL3.5-4B & 54.6 & 48.5 & 29.7 & 46.4 & 53.3 & 51.4 & 35.4 & 50.7 \\
Qwen3-VL-4B & 55.0 & 47.9 & 25.3 & 46.1 & 42.4 & 38.1 & 17.5 & 50.9 \\
PLM-3B & 47.4 & 46.3 & 21.0 & 47.3 & 39.6 & 39.8 & 20.3 & 46.9 \\
V-JEPA caps+QA (2B, 8k) & 56.6 & 47.9 & 32.3 & 40.1 & 50.7 & 50.6 & 34.5 & 48.1 \\
SigLIP caps+QA (2B, 8k) & 43.5 & 43.0 & 19.4 & 49.4 & 35.6 & 38.5 & 14.8 & 62.8 \\
SigLIP +CF (2B, 8k) & 43.6 & 40.4 & 19.7 & 48.5 & 37.2 & 38.1 & 14.3 & 57.0 \\
\rowcolor{ours}V-JEPA +CF (2B, 8k) & 59.8 & 52.2 & 36.8 & 32.7 & 52.3 & 52.8 & 38.0 & 41.4 \\
\rowcolor{ours}Ours-2B-Q & 65.5 & 54.7 & 40.7 & 31.6 & 57.3 & 58.2 & 42.7 & 37.1 \\
\rowcolor{ours}Ours-1B-L & 63.2 & 53.6 & 41.0 & 31.3 & 55.1 & 59.3 & 42.9 & 36.0 \\
\rowcolor{ours}Ours-4B-Q & 70.0 & 57.9 & 47.7 & 27.3 & 59.4 & 61.2 & 50.4 & 33.0 \\
\bottomrule
\end{tabular}
\caption{Counterfactual consistency in full. Items qualify automatically when the correct option, with the transformation's word swaps applied (left$\leftrightarrow$right, towards$\leftrightarrow$away, open$\leftrightarrow$close, before$\leftrightarrow$after, \dots), equals another option exactly. Orig: accuracy on the original video; follows: the transformed answer is the swapped option; both: both correct; stuck: same answer for both videos.}
\label{tab:app_cf}
\end{table*}

\FloatBarrier
\section{Internal measurements}
\label{app:internal}
All measurements use clean pairs and the exact inputs and answer position of evaluation, with no generation (2B, 8k steps).
Share is $\mathbb{E}d_s^2/(\mathbb{E}d_s^2+\mathbb{E}m^2)$ of the answer log-odds over a pair set (\cref{app:theory}); stuck is the fraction of pairs given the same answer for both videos. Leak is the variance of the content along the answer direction $u=\mathrm{norm}_w\odot(W_P-W_Q)$, relative to the change.

\begin{table}[!htbp]
\centering\footnotesize\setlength{\tabcolsep}{2.2pt}
{\begin{tabular}{lrrr}
\toprule
Pair set: share / stuck / leak & SigLIP +CF & V-JEPA caps+QA & \cellcolor{ours}V-JEPA +CF \\
\midrule
Unseen speed ratios & .065 / 81.7\% / 44.8 & .022 / 89.4\% / 52.7 & \cellcolor{ours}.983 / 1.1\% / 0.08 \\
Speed, new source (Kinetics-700) & .611 / 73.2\% / 1.24 & .081 / 90.8\% / 5.3 & \cellcolor{ours}.943 / 6.3\% / 0.29 \\
Accel./decel.\ (+reversal) & .482 / 33.1\% / 2.0 & .081 / 79.5\% / 9.1 & \cellcolor{ours}.745 / 5.4\% / 0.83 \\
Direction combinations & .918 / 51.0\% / .087 & .900 / 9.2\% / .095 & \cellcolor{ours}.980 / 2.7\% / .021 \\
\midrule
$\kappa$ / proto 8 at layer 18 & & & \\
Unseen speed ratios & .008 / 50.1 & .035 / 55.2 & \cellcolor{ours}.648 / 99.3 \\
Speed, new source & .105 / 42.0 & .056 / 47.1 & \cellcolor{ours}.324 / 63.1 \\
Accel./decel. & .229 / 78.9 & .100 / 58.6 & \cellcolor{ours}.207 / 89.4 \\
Direction combinations & .076 / 51.0 & .288 / 93.1 & \cellcolor{ours}.300 / 93.8 \\
Direction, bridge input & .0006 / 49.9 & .361 / 90.5 & \cellcolor{ours}.330 / 90.6 \\
\midrule
Parallelogram cosine (layer 27) & & & \\
mirror effect & $-$.293 & $-$.894 & \cellcolor{ours}$-$.938 \\
reversal effect & +.252 & $-$.877 & \cellcolor{ours}$-$.888 \\
mirror effect, bridge input & +.777 & $-$.242 & \cellcolor{ours}$-$.235 \\
Composition acc.\ / original & 69.9 / 65.1 & 93.8 / 94.5 & \cellcolor{ours}95.2 / 98.6 \\
\bottomrule
\end{tabular}}
\caption{Internal measurements. Parallelogram: $\cos(h_{\text{plain}}-h_{\text{mirror}},\,h_{\text{rev}}-h_{\text{mirror+rev}})$; $-1$ means mirror and reversal flip the same subspace (\cref{thm:a}); clips paired with other clips give about 0.03.}
\label{tab:app_internal}
\end{table}

\paragraph{Removing change subspaces.} \Cref{tab:app_ablate} removes the change directions measured on frozen features from every visual token at inference, with random and content subspaces as controls.
\begin{table*}[!htbp]
\centering\footnotesize\setlength{\tabcolsep}{3.5pt}
\begin{tabular}{llrrrrrrrrrrrr}
\toprule
& & & & \multicolumn{3}{c}{held-out speed} & \multicolumn{2}{c}{held-out direction} & & \multicolumn{2}{c}{natural, dynamics} & \multicolumn{2}{c}{natural, content} \\
\cmidrule(lr){5-7}\cmidrule(lr){8-9}\cmidrule(lr){11-12}\cmidrule(lr){13-14}
Model & Removed subspace & dims & energy & ratio & dom. & comb. & dom. & comb. & camera & TOM & TC & NQA & PT \\
\midrule
\rowcolor{ours}Ours-2B-Q & none & 0 & 0 & 98.7 & 98.5 & 98.9 & 75.7 & 98.5 & 99.8 & 50.9 & 73.5 & 74.0 & 75.3 \\
\rowcolor{ours} & speed & 8 & 16.6 & 91.5 & 77.8 & 65.7 & 75.2 & 98.5 & 98.8 & 46.4 & 70.1 & 73.2 & 73.8 \\
\rowcolor{ours} & random, energy-matched & 24 & 16.2 & 96.7 & 95.5 & 97.2 & 74.8 & 98.1 & 99.2 & 49.9 & 71.7 & 73.2 & 74.3 \\
\rowcolor{ours} & reversal & 8 & 6.0 & 96.9 & 97.8 & 98.9 & 72.4 & 97.9 & 99.4 & 50.3 & 71.6 & 74.0 & 75.2 \\
\rowcolor{ours} & mirror & 8 & 3.0 & 98.7 & 98.7 & 99.4 & 74.4 & 97.6 & 99.6 & 50.0 & 74.2 & 74.1 & 75.2 \\
\rowcolor{ours} & speed + reversal + mirror & 24 & 21.5 & 81.3 & 74.7 & 63.0 & 71.9 & 96.9 & 98.4 & 46.0 & 68.9 & 73.1 & 73.8 \\
\rowcolor{ours} & random & 24 & 2.4 & 98.1 & 98.3 & 99.1 & 75.4 & 97.9 & 99.8 & 50.6 & 72.5 & 74.0 & 74.7 \\
\rowcolor{ours} & top content PCs & 24 & 43.6 & 95.0 & 85.0 & 84.3 & 75.6 & 97.4 & 96.5 & 44.8 & 68.1 & 72.4 & 72.0 \\
\midrule
V-JEPA caps+QA & none & 0 & 0 & 33.3 & 44.3 & 49.1 & 69.8 & 97.1 & 92.4 & 37.5 & 66.5 & 74.4 & 74.2 \\
 & speed + reversal + mirror & 24 & -- & 33.5 & 43.5 & 50.2 & 59.8 & 89.2 & 76.3 & 33.6 & 64.9 & 73.7 & 72.6 \\
\bottomrule
\end{tabular}
\caption{Removing change subspaces at the bridge input (final 2B checkpoints, no retraining). The subspaces are estimated once in the equalised space of Ours-2B-Q from 400 SSv2 \emph{training} clips: the top 8 directions of the token-averaged response to playing at twice the speed, to reversal and to mirroring (on the other 20\% of clips they capture 46\%, 57\% and 74\% of that response); every visual token is projected off the subspace. Energy: share of token energy removed (\%). Energy-matched random: a random 24-dimensional subspace inside a band of content principal components with the same removed energy as the speed subspace. Speed removal takes the unseen speed combinations from 98.9 to 65.7 and TOMATO from 50.9 to 46.4, while the energy-matched random subspace leaves them at 97.2 and 49.9; content benchmarks move by at most 1.5 in both. The model trained without counterfactual rows is at chance on speed with or without the subspace.}
\label{tab:app_ablate}
\end{table*}

\paragraph{Training rows themselves.}
On the training distribution, held-out rows of each construction have answer loss as follows (without $\to$ with CF):
\begin{itemize}[leftmargin=*,itemsep=0pt,topsep=1pt]
\item \emph{relative speed:} V-JEPA $0.23\to0.02$, while SigLIP stays at $0.22\to0.22$, failing even on the training construction;
\item \emph{direction and camera:} $0.10\to0.03$ and $0.23\to0.12$;
\item \emph{order, position and count:} V-JEPA $0.16\to0.07$ and SigLIP $0.18\to0.07$, the same for both encoders, as \cref{thm:a} predicts for signals in token positions.
\end{itemize}

\FloatBarrier
\section{Explicit pairs versus unpaired rows}
\label{app:pairs}
\paragraph{Construction.} All counterfactual rows form one pool. A \emph{scene} is the footage a row shows (an SSv2 or Kinetics clip, a CLEVRER or simulation video, or one-second bins of a long video); two rows share a scene if their footage overlaps, and two rows are \emph{counterparts} if they share a scene, belong to the same question family and have different answer classes.
\emph{Unpaired:} rows are taken in a seeded random order, always for the (source, answer class) furthest below its original share, and a row is kept only if none of its scenes is used yet, so every scene appears in at most one row.
\emph{Paired:} connected groups of counterparts are taken whole, in random order, while every (source, answer class) stays within the unpaired arm's count; rows without counterparts fill the rest.
Both arms thus have the same number of rows per source and per answer class (217k in total, 43\% of the original) and the same per-source weights; they differ in which rows are drawn. Counterparts are counted here by shared footage, also across sources (35.9\% of the full pool's rows have one, against 29.4\% within a source in \cref{tab:app_data}); in the paired arm 29.4\% of rows have their counterpart (66.8\% for direction, 100\% for physics), in the unpaired arm none do.
\begin{itemize}[leftmargin=*,itemsep=0pt,topsep=1pt]
\item \emph{Twelve-score averages:} 61.56 and 60.27 (8k steps), against 61.18 for the full mixture.
\item \emph{Ability differences (paired $-$ unpaired):} state change $+4.5$, reversal/antonym $+4.6$, TemporalBench $+2.8$, speed $+3.8$ and count $-3.1$.
\item \emph{Held-out sets:} both average 92.1.
\item \emph{Unseen speed-ratio pairs:} both answers are correct for 96.7\% against 84.4\%, with dynamic shares 0.978 / 0.976.
\end{itemize}
The comparison is between two ways of selecting rows at the same per-source and per-class budget; the unpaired arm covers 22\% more distinct videos (194k against 159k), so coverage favors it. The dynamic shares are nearly equal, so the paired arm's gain is in consistency across a pair rather than in a larger differential share. The speed sources contain no explicit pairs in either arm (\cref{tab:app_data}), so the speed gains of the full method come from answer-balanced construction.
At equal error weights the gradient's content term cancels in expectation in both models and within each scene with explicit pairs, which pays off where scene and answer correlate in natural video.

\FloatBarrier
\section{Scope and limitations}
\label{app:limits}
The aim of this work is a light video-language model that reads the dynamics a predictive world-model encoder already holds, together with the analysis that explains when and why this works. The study is scoped to that aim.

\paragraph{Short clips.}
All models read 32 frames per clip, and all benchmarks pose questions about clips of seconds to a few minutes. Hour-long video understanding, which needs memory, retrieval across many clips and long-context training data, is a different problem, and training for it at our model sizes and budget would be impractical. The compact 1,024-token clip representation is a natural building block for it, which we leave to future work.

\paragraph{Image question answering.}
We do not target image question answering (documents, charts, OCR, general image QA). V-JEPA is trained only to predict masked regions of video in latent space, never to align with or read text, so text recognition lies outside what the encoder was built for; performance there is driven by very large image--text corpora. For the same reason, open models trained on 2 to 38 times more data keep an edge on knowledge-heavy video benchmarks such as NExT-QA and PerceptionTest (\cref{tab:main}).

\paragraph{Data and compute scale.}
Each model is trained once on 4.5M public samples, with at most 650 H100-hours. Training and evaluating at the scale of general-purpose VLMs, with hundreds of millions of samples or trillions of tokens including long video and large image QA collections, is outside both the purpose of this study and the resources available to it. The results indicate where such scale would pay off: the counterfactual rows are generated from public videos without new manual annotation, make up 6\% of the mixture, and, for the encoders compared here, their benefit is anticipated by measurements on the frozen encoder.

\end{document}